\documentclass{article}

\usepackage{microtype}
\usepackage{graphicx}
\usepackage{subfigure}
\usepackage{booktabs} 
\usepackage{microtype}
\usepackage{graphicx}
\usepackage{subfigure}
\usepackage{booktabs} 
\usepackage{lipsum}
\usepackage{xcolor}
\usepackage{url}
\usepackage{bbding}

\usepackage{epsfig}
\usepackage{array}
\usepackage{multirow}
\usepackage{epstopdf}
\usepackage{tikz}
\usepackage{relsize}
\newcommand{\E}{\mathbb{E}}
\newcommand{\e}{\Sigma}
\usepackage{hyperref}
\definecolor{mygreen}{rgb}{0.0, 0.5, 0.0}
\definecolor{winered}{rgb}{0.8,0,0}
\definecolor{myblue}{rgb}{0,0,0.8}
\hypersetup{
    colorlinks=true,
    linkcolor={winered},
    citecolor={myblue}
}

\usepackage[accepted]{icml2024}

\usepackage{amsmath}
\usepackage{amssymb}
\usepackage{mathtools}
\usepackage{amsthm}
\usepackage{tcolorbox}
\usepackage[capitalize,noabbrev]{cleveref}

\theoremstyle{plain}
\newtheorem{theorem}{Theorem}[section]

\newtheorem{lemma}[theorem]{Lemma}
\newtheorem{corollary}[theorem]{Corollary}
\theoremstyle{definition}

\newtheorem{assume}[theorem]{Assumption}
\theoremstyle{remark}

\newtheorem{prop}{Proposition}

\usepackage[textsize=tiny]{todonotes}

\usepackage{hyperref}

\usepackage[accepted]{icml2024}

\usepackage{amsmath}
\usepackage{amssymb}
\usepackage{mathtools}
\usepackage{amsthm}

\usepackage[capitalize,noabbrev]{cleveref}

\usepackage[textsize=tiny]{todonotes}

\icmltitlerunning{Momentum for the Win:  Collaborative Federated Reinforcement Learning across Heterogeneous Environments}

\begin{document}

\twocolumn[
\icmltitle{Momentum for the Win:  Collaborative Federated Reinforcement Learning across Heterogeneous Environments}



\icmlsetsymbol{equal}{*}

\begin{icmlauthorlist}
\icmlauthor{Han Wang}{sch}
\icmlauthor{Sihong He}{yyy}
\icmlauthor{Zhili Zhang}{yyy}
\icmlauthor{Fei Miao}{yyy}
\icmlauthor{James Anderson}{sch}
\end{icmlauthorlist}
\icmlaffiliation{sch}{Department of Electrical Engineering, Columbia University, New York, USA.}
\icmlaffiliation{yyy}{School of Computing, University of Connecticut, Storrs, USA}

\icmlcorrespondingauthor{Han Wang}{hw2786@columbia.edu}

\icmlkeywords{Machine Learning, ICML}

\vskip 0.3in
]



\printAffiliationsAndNotice{}  

\begin{abstract}
We explore a Federated Reinforcement Learning (FRL) problem where $N$ agents collaboratively learn a common policy without sharing their trajectory data. To date, existing FRL work has primarily focused on agents operating in the same or ``similar" environments. In contrast, our problem setup allows for arbitrarily large levels of environment heterogeneity. To obtain the optimal policy which maximizes the average performance across all  \emph{potentially completely different} environments, we propose two algorithms: \textsc{FedSVRPG-M} and \textsc{FedHAPG-M}. In contrast to existing results, we demonstrate that both \textsc{FedSVRPG-M} and \textsc{FedHAPG-M}, both of which leverage momentum mechanisms, can exactly converge to a stationary point of the average performance function, regardless of the magnitude of environment heterogeneity. Furthermore, by incorporating the benefits of variance-reduction techniques or Hessian approximation, both algorithms achieve state-of-the-art convergence results, characterized by a sample complexity of  $\mathcal{O}\left(\epsilon^{-\frac{3}{2}}/N\right)$. Notably, our algorithms enjoy linear convergence speedups with respect to the number of agents, highlighting the benefit of collaboration among agents in finding a common policy.
\end{abstract}

\section{Introduction}
Recently, there has been a lot of interest  applying Federated Learning (FL) algorithms to reinforcement learning (RL) problems in order to solve  complex sequential decision-making tasks~\citep{qi2021federated,jin2022federated,liu2019lifelong}. Federated reinforcement learning (FRL) has been widely applied as it provides the following advantages: First, FRL protects each agent's privacy by only allowing the model to be shared between the server and agent, while keeping the raw data localized. Secondly, by sharing the  model with the server, FRL can reduce the sample complexity and produce a better policy than if each agent learns individually with its own limited data. However, existing work in the FRL framework is  limited to either multiple agents interacting with the same environment~\citep{fan2021fault,khodadadian2022federated} \emph{or} multiple agents with distinct, \emph{yet similar} environments~\citep{jin2022federated,xie2023fedkl,wang2023federated}. It remains an open problem to formally characterize how FRL performs when multiple agents from completely different environments, i.e., with arbitrarily large heterogeneity levels, are allowed to collaborate.  In this work, we provide an answer to the following question: \emph{what is the best achievable sample complexity when considering severely heterogeneous environments}?


We focus on developing FRL algorithms that compute an optimal universal policy that ensures uniformly good performance for 
$N$ agents, despite their operation in disparate environments. The motivation for a shared policy stems from practical applications necessitating uniform approaches for distinct agents. For instance, Spotify, a leading audio streaming company, intends to design a uniform pricing plan that suits the listening habits of all users. Given the substantial variations in listening habits among users, establishing a pricing strategy that aligns with the preferences of all users is of great importance. Similarly, autonomous vehicles navigating diverse settings like urban streets, rural areas, and highways must adapt to varied challenges. A uniform policy that adjusts to this environmental heterogeneity ensures consistent, safe decision-making across all terrains, highlighting the need for robust algorithms capable of handling dynamic driving conditions efficiently. Moreover, a universally optimal policy could serve as a foundational model that can be individually fine-tuned, a concept that has gained a lot of attention in meta- and few-shot RL research \citep{finn2017model,yu2020meta}. This approach underscores the broader necessity of designing a uniform and adaptable policy for heterogeneous settings.

In this work, the environment heterogeneity refers to the fact that each agent has a different reward function, state transition kernel, or initial state distribution, while they share common state and action spaces. Notably, compared with the existing work~\citep{jin2022federated, wang2023federated}, we do not assume that all the environments are similar, i.e., environmental heterogeneity does not need to be bounded by  small constants. Instead, we consider a more general setting where the magnitude of heterogeneity can be arbitrary. With this setup, we aim to answer the following question:
\begin{quote}
  \textit{Is it possible to design a provably efficient FRL algorithm which can accommodate arbitrary levels  of environmental heterogeneity among agents?}
\end{quote}
We answer this question affirmatively. Our main contributions are listed below.

 $\bullet$  \textbf{New momentum-powered federated reinforcement learning algorithms:} 
 We propose two new algorithms \textsc{FedSVRPG-M} and \textsc{FedHAPG-M} for solving heterogeneous FRL problems (formally specified in Eq.~\eqref{eq:main_formulation}). Leveraging momentum, we prove that our algorithms, even with constant local step-sizes, converge to the exact stationary point of the heterogeneous FRL problem, \emph{regardless of the magnitude of environment heterogeneity}. This stands in contrast to the state-of-the-art work, which only show convergence to a ball around the stationary point whose radius depends on the environmental heterogeneity levels. Importantly, our results hold even when different notions of environment heterogeneity are considered such as the heterogeneity in Markov decision processes (MDPs) or policy advantage heterogeneity~\citep{xie2023fedkl}. 


 $\bullet$ \textbf{State-of-the-art convergence rates:} 
 By integrating variance-reduction techniques and curvature information into the policy gradient estimation, our algorithms achieve  sample-efficiency improvement over prior work~\citep{fan2021fault}. In particular, we reduce the sample complexity from $
\mathcal{O}\left( \epsilon^{-\frac{5}{3}}/N^{\frac{2}{3}}\right)$ to $\mathcal{O}\left(\epsilon^{-\frac{3}{2}}/N\right)
$ when finding the $\epsilon$-approximate first order stationary point\footnote{Finding a parameter $\theta$ such that $\|\nabla J(\theta)\|^2 \leq \epsilon$, where $J$ is defined in Eq.~\eqref{eq:main_formulation}. Note that in work such as~\cite{shen2019hessian,fatkhullin2023stochastic}, the notion $\|\nabla J(\theta)\|^2 \leq \epsilon^2$ is applied instead. } ($\epsilon$-FOSP)~\citep{nesterov2003introductory}. When only a single agent is included, i.e., $N=1$, our results align with the best known sample complexity of ${\mathcal{O}}\left(\varepsilon^{-\frac{3}{2}}\right)$ from~\citet{fatkhullin2023stochastic}.

 $\bullet$  \textbf{Practical algorithm structures:} 
 Our algorithms are easy to implement because: (1) \emph{Constant local step-sizes}. This feature reduces the amount of  algorithm tuning. In contrast, many FL optimization algorithms~\citep{karimireddy2020scaffold,wang2019slowmo,yang2021achieving} require diminishing local step-sizes preset according to complex schedules in order to counteract the effects of heterogeneity. (2) \emph{Sampling one trajectory per local iteration}. This means our algorithms can address the challenge of poor sample efficiency in RL. Unlike existing variance-reduced policy gradient (PG) algorithms for the single agent setting~\citep{papini2018stochastic,xu2019sample,gargiani2022page}, our approach avoids the need for large batch sizes during certain iterations. (3) \emph{Accommodating multiple local updates}. With this feature, our algorithms become more suitable for real-world applications, where communication latency causes serious bottlenecks.
 
 $\bullet$  \textbf{Linear speedup:} 
Analysis of \textsc{FedSVRPG-M} and \textsc{FedHAPG-M} shows that they can converge $N$-times faster than the scenario where each agent learns a policy on its own. Essentially, by adopting the FL approach, the sample complexity of our algorithms can be linearly scaled by the number of agents $N$, i.e., collaboration always helps. To our knowledge, \emph{we are the first to achieve a linear speedup for finding a stationary point of FRL problems using policy-based methods}. Importantly, the linear speedup is established even when considering multiple local updates and without making any assumptions about environment heterogeneity. Compared to prior work, our result outperforms that of ~\citet{jin2022federated,fan2021fault}, which at best achieves sublinear speedup, see Table~\ref{tab:1}.

\begin{table*}[tb]
\caption{Comparision of the results for policy-based methods in FRL. LU and HETER denote the multiple local updates and environment heterogeneity, respectively.}
\centering
  \begin{tabular}{lllll}
  \\
 \bf{ALGORITHM}  &  \bf{CONVERGENCE}  & \bf{SPEEDUP}  & \bf{LU} & \bf{HETER} \\
\hline\\ \textsc{PAvg} \citep{jin2022federated}  &   finite but inexact & No speedup & \Checkmark & \Checkmark \\
  \textsc{FedKL} \citep{xie2023fedkl}  & asymptotic &No speedup & \XSolidBrush & \Checkmark \\
\textsc{FedPG-BR} \cite{fan2021fault}  &  finite and exact & \text{Sublinear}: $N^{
\frac{2}{3}
}$ & \XSolidBrush & \XSolidBrush \\
\textsc{FAPI} \citep{xie2023client}& asymptotic and inexact & No speedup& \XSolidBrush& \Checkmark\\
\textsc{FedSVRPG-M} (\textcolor{red}{Ours}) &  finite and exact  & Linear: $N$ & \Checkmark & \Checkmark \\
\textsc{FedHAPG-M} (\textcolor{red}{Ours})  & finite and exact & \text{Linear}: $N$ & \Checkmark & \Checkmark \\
\hline
\end{tabular}
  
  \label{tab:1}
\end{table*}

\section{Related Work}
\paragraph{Federated RL} A comprehensive overview of techniques and open problems in FRL was offered by~\citet{qi2021federated}. Much of the work in FRL has focused on developing federated versions of value-based methods~\cite{khodadadian2022federated,woo2023blessing,wang2023federated}. Notably, 
 \citet{khodadadian2022federated} and \citet{woo2023blessing} established the benefits of FL in terms of linear speedup, assuming all agents operate in \emph{identical} environment. Wang et al. \citet{wang2023federated} introduced the \textsc{FedTD}(0) algorithm to address the FRL problem with \emph{distinct yet similar} environments demonstrated linear speed up was achievable. On the other hand, \citet{zhang2024finite} proposed the \textsc{FedSARSA} algorithm to solve the on-policy FRL problem, but it is applicable only in similar environments. Another major area of FRL research studies federated policy-based algorithms~\citep{jin2022federated,xie2023fedkl,fan2021fault, wang2023model, lan2023improved}. However,  \citet{fan2021fault} only consider uniform environments and only one local update step. While \citeauthor{xie2023fedkl} explored diverse environments, they only showed an asymptotic convergence. Most relevant to our work, \citet{jin2022federated} studied \emph{heterogeneous} environments. Nevertheless, the algorithms from \citet{jin2022federated} were saddled with a non-vanishing convergence error. This non-vanishing error depended on the  environmental heterogeneity levels. Note that none of these papers investigated the FRL problems with \emph{arbitrary environment heterogeneity}. To bridge this gap, our proposed algorithms,  \textsc{FedSVRPG-M} and \textsc{FedHAPG-M}, utilize policy-based techniques and can converge  \emph{exactly}. See Table~\ref{tab:1} for a comparison of our results with the existing work in FRL policy-based methods.

\section{Preliminaries}
\textbf{Federated Learning.} Federated learning (FL) is a machine learning approach where a model is trained across multiple clients. Each client runs several iterations of a learning algorithm on its own dataset. Periodically, clients send their local models to the server. The server  aggregates the models and then broadcasts the resulting model to all clients and the process repeats.  By performing multiple local updates with its own data, FL can substantially reduce communication costs. Our proposed algorithms align with the structure of standard FL algorithms such as \textsc{FedAVG}~\citep{mcmahan2017communication}: an agent performs multiple local updates (using SGD) between two communication rounds. Nonetheless, such local updates will introduce \textit{``client-drift"} problems~\citep{karimireddy2020scaffold,charles2021convergence,wang2022fedadmm}, presenting a key challenge in FL regarding the trade-off between communication cost and model accuracy. Additionally, handling data that is not identically distributed across devices, affecting both data modeling and convergence analysis, presents another challenge.  These challenges are further amplified in the context of FRL. 

\subsection{Centralized Reinforcement Learning}
A centralized reinforcement learning task\footnote{To distinguish from the federated setting, we refer to the single-agent case as  centralized RL or when it's clear from context, simply reinforcement learning.} is generally modeled as a discrete-time Markov Decision Process (MDP): $\mathcal{M}=\{\mathcal{S}, \mathcal{A}, \mathcal{P}, \mathcal{R}, \gamma, \rho\}$, where $\mathcal{S}$ is the state space, $\mathcal{A}$ is the action space and $\rho$ denotes the initial state distribution. Here, $\mathcal{P}\left(s^{\prime} \mid s, a\right)$  denotes the probability that the agent transitions from the state $s$ to $s^{\prime}$ when taking the action $a \in \mathcal{A}$. The discount factor is $\gamma \in (0,1)$, and $\mathcal{R}(s, a): \mathcal{S} \times \mathcal{A} \rightarrow[0, R_{\max}]$ is the reward function for taking action $a$ at state $s$ for some constant $R_{\max}>0$. A policy $\pi: \mathcal{S} \rightarrow \Delta(\mathcal{A})$ is a mapping from the state space $\mathcal{S}$ to the probability distribution over the action space $\mathcal{A}$. 

Under any stationary policy, the agent can collect a trajectory $\tau \triangleq \left\{s_0, a_0, s_1, a_1, \ldots, s_{H-1}, a_{H-1}, s_H\right\}$, which is the collection of state-action pairs, where $H$ is the maximum length of all trajectories. Once a trajectory $\tau$ is obtained, a cumulative discounted reward can be observed; 
$
\mathcal{R}(\tau)\triangleq \sum_{h=0}^{H-1} \gamma^h \mathcal{R}\left(s_h, a_h\right).
$

\subsection{Policy Gradients}
Given finite state and action spaces, the policy $\pi(a|s)$ can be stored in a $|\mathcal{S}| \times |\mathcal{A}|$ table. However, in practice, both the state and action spaces are large and the tabular approach becomes intractable. Alternatively, the policy  is parameterized by an unknown parameter $\theta \in \mathbb{R}^d,$ the resulting policy is denoted by $\pi_{\theta}$. Given the initial distribution $\rho$,  $p(\tau \mid {\theta})$ denotes the probability distribution over trajectory $\tau$, which can be calculated as
\begin{align*}\label{eq:trajectory_distribution}
p(\tau \mid {\theta})=\rho\left(s_0\right) \prod_{h=0}^{H-1} \pi_{{\theta}}\left(a_h \mid s_h\right) \mathcal{P}\left(s_{h+1} \mid s_h, a_h\right).
\end{align*}
The goal of RL is to find the optimal policy parameter $\theta$ that maximizes the expected discounted trajectory reward:
\begin{equation}\label{eq:single_objective}
\max _{\theta \in \mathbb{R}^d} J(\theta)\triangleq \mathbb{E}_{\tau \sim p(\tau \mid \theta)}[\mathcal{R}(\tau)]=\int \mathcal{R}(\tau) p(\tau \mid \theta) d \tau.
\end{equation}
Note that the underlying distribution $p$ in Eq.~\eqref{eq:single_objective} depends on the variable $\theta$ which varies through the whole optimization procedure. This property, referred to as \textit{non-obliviousness}, highlights a unique challenge in RL and creates a notable distinction from supervised learning problems, where the distribution $p$ is stationary.

To deal with the \textit{non-oblivious} and \textit{non-convex}  problem \eqref{eq:single_objective}, a standard approach is to use the policy gradient (PG) method~\citep{williams1992simple,sutton1999policy}. PG takes the first-order derivative of the objective~\eqref{eq:single_objective} where $\nabla J(\theta)$ can be expressed as
\begin{align*}
\int \mathcal{R}(\tau) \nabla p(\tau \mid \theta) d \tau  =\mathbb{E}_{\tau \sim p(\tau \mid \theta)}[\nabla \log p(\tau \mid \theta) \mathcal{R}(\tau)].
\end{align*}
Then, the policy $\theta$ can be optimized by running gradient ascent-based algorithms. However, since the distribution $ p(\tau \mid \theta)$ is unknown, it is impossible to calculate the full gradient. To address this issue, stochastic gradient ascent is typically used, producing a sequence of the form: 
$$\theta \leftarrow \theta +\eta \cdot \frac{1}{B}\sum_{i=1}^{B}g(\tau_i \mid \theta)$$
where $\eta >0$ denotes the stepsize, $B$ is the number of trajectories, and $g(\tau_i \mid \theta)$ is an estimate of the full gradient $\nabla J(\theta)$ using the trajectory $\tau_i$. The most common unbiased estimators of PG are REINFORCE~\citep{williams1992simple} and GPOMDP~\citep{baxter2001infinite}. In this paper, $g(\tau\mid \theta)$ is defined as $$g(\tau \mid \theta)=\sum_{t=0}^{H-1}\left(\sum_{h=t}^{H-1} \gamma^h \mathcal{R}\left(s_h, a_h\right)\right) \nabla \log \pi_\theta\left(a_t \mid s_t\right).$$

\paragraph{Importance Sampling} Since problem~\ref{eq:single_objective} is \textit{non-oblivious}, we have $\mathbb{E}_{\tau \sim p(\tau \mid \theta)}\left[g(\tau \mid \theta)-g\left(\tau \mid \theta^{\prime}\right)\right] \neq \nabla J(\theta)-\nabla J\left(\theta^{\prime}\right)$. To address this issue of distribution shift, we introduce an importance sampling (IS) weight, denoted by
\begin{equation}\label{eq:IS_weight}
w\left(\tau \mid \theta^{\prime}, \theta\right) \triangleq \frac{p\left(\tau \mid \theta^{\prime}\right)}{p(\tau \mid \theta)}=\prod_{h=0}^{H-1} \frac{\pi_{\theta^{\prime}}\left(a_h \mid s_h\right)}{\pi_\theta\left(a_h \mid s_h\right)}.
\end{equation}
With the definition of the IS weight, we can ensure that $\mathbb{E}_{\tau \sim p(\tau \mid \theta)}\left[g(\tau \mid \theta)-w\left(\tau \mid \theta^{\prime}, \theta\right) g\left(\tau \mid \theta^{\prime}\right)\right]=\nabla J(\theta)-\nabla J\left(\theta^{\prime}\right).$

\section{Problem Formulation}
We are now ready to characterize heterogeneity in our $N$-agent FRL problem. Environmental heterogeneity is modeled by allowing each agent to have its own state transition kernel $\mathcal{P}^{(i)}$, reward function $\mathcal{R}^{(i)}$, or the initial state distribution $\rho^{(i)}$. However, all agents share the same state and action space. These environments are characterized by the  MDPs, $\mathcal{M}_i=\left\langle\mathcal{S}, \mathcal{A}, \mathcal{R}^{(i)}, \mathcal{P}^{(i)}, \gamma, \rho^{(i)}\right\rangle$, for $i=1, \cdots, N$.

The objective of FRL is to enable $N$ agents to collaboratively learn a common policy function or a value function that uniformly performs well across all environments. To preserve privacy,  agents are not allowed to exchange their raw observations (i.e., their rewards, states, or actions). In particular, we consider solving the following optimization problem: 
\begin{align}\label{eq:main_formulation}
\max _{\theta}  &  \quad \left\{J ( \theta )\triangleq\frac{1}{N}\sum_{i=1}^N J_i (\theta) \right\}\notag\\
 \text{where} & \quad  J_i (\theta)\triangleq \mathbb { E } \left[\sum_{h=0}^{H-1} \gamma^h \mathcal{R}^{(i)}\left(s_h, a_h\right) \mid s_0 \sim \rho^{(i)},\right.\notag\\
 &\left.a_h \sim \pi_{\theta}\left(\cdot \mid s_h\right), s_{h+1} \sim \mathcal{P}^{(i)}\left(\cdot \mid s_h, a_h\right)\right].
\end{align}
\paragraph{Objective.} For solving the optimization problem \eqref{eq:main_formulation}, we aim to find the $\epsilon$-FOSP, i.e., a parameter $\theta$ such that $\|\nabla J(\theta)\|^2 \leq \varepsilon.$ There exists work that leverages the ``gradient domination" condition \citep{agarwal2020optimality,liu2020improved,ding2021beyond,fatkhullin2023stochastic} for finding a global optimal policy in the centralized RL setting. The gradient domination condition is useful as it guarantees that every stationary policy is globally optimal. However, as shown in Zeng et al.~\yrcite{zeng2021decentralized}, we cannot expect this condition to hold in general for  FL or multi-agent problems. Specifically, even if a single performance function, $J_i(\theta)$, satisfies the ``gradient domination" condition, the average function $J(\theta)=\frac{1}{N}\sum_{i=1}^N J_i(\theta)$ might not. \citet{zeng2021decentralized} resolved this issue by introducing strong assumptions into the problem. For instance, Assumption 2 in their paper requires that the joint states between the environments are equally explored, which is difficult to verify in real-world applications. 

\textbf{Difference in the problem setup.} Our setting is more general than existing work~\citep{jin2022federated,wang2023federated}. In our work, each MDP can have a distinct initial state distribution, a feature not addressed in~\citeauthor{jin2022federated}. Furthermore, our framework does not require the bounded heterogeneity assumption of~\citeauthor{wang2023federated} and thus can handle arbitrary environment heterogeneity. 

\section{Algorithms} 
To solve problem~\eqref{eq:main_formulation}, we present two federated momentum-based algorithms: \textsc{FedSVRPG-M} and \textsc{FedHAPG-M}. \textsc{FedSVRPG-M} is based on a variance reduction method, while \textsc{FedHAPG-M} leverages a fast Hessian-aided technique. Since \textsc{FedSVRPG-M} only uses the first-order information (gradient), it is computationally cheaper than \textsc{FedHAPG-M}, which aims to approximate second-order information (Hessians). Conversely, \textsc{FedHAPG-M}, with its use of second-order information, is more robust than \textsc{FedSVRPG-M}.

In the centralized RL setting,  momentum-based PG methods~\citep{yuan2020stochastic,huang2020momentum} are proposed to reduce the variance of stochastic gradients. In contrast, our algorithms integrate momentum within a federated context, achieving dual benefits: it not only accelerates the convergence and stabilizes oscillations, but also mitigates the impact of environment heterogeneity. Consequently, our algorithms can exactly converge to the $\epsilon$-FOSP of problem~\eqref{eq:main_formulation}, no matter how large the environment heterogeneity is. This represents a significant improvement upon~\citep{jin2022federated,xie2023client}, which only show the convergence to the neighborhood around the stationary point of problem. The size of the neighborhood in their papers is determined by the environment heterogeneity.


\subsection{\textsc{FEDSVRPG-M}}
We now describe the federated stochastic variance-reduced PG with momentum algorithm (\textsc{FedSVRPG-M} for short). We outline its
steps in Algorithm~\ref{algFedRL}.  

\textsc{FEDSVRPG-M} initializes all agents and the server with a common model $\theta_{0}.$ In Algorithm~\ref{algFedRL}, we use the superscript $(i)$ to index the $i$-th agent and the subscript $r$ and $k$ to denote the $r$-th communication round and $k$-th local iteration. In each communication round $r$, each agent $i \in [N]$ is initiated from a common model $\theta_r$ and samples a single trajectory from its own environment to perform $K$ local iterations. At each local iteration $k$, instead of using PG, \textsc{FedSVRPG-M} uses the following momentum-based variance-reduced stochastic PG estimator:
\begin{align}\label{eq:gradient_local}
u_{r,k}^{(i)}&={\beta} g_i\left(\tau_{r,k}^{(i)}\mid \theta_{r,k}^{(i)}\right)+{\left(1-\beta\right)}\left[u_r+g_i\left(\tau_{r,k}^{(i)}\mid \theta_{r,k}^{(i)}\right)\right.\notag\\
&\left.- w^{(i)}\left(\tau_{r,k}^{(i)} \mid \theta_{r-1}, \theta_{r,k}^{(i)}\right) g_i\left(\tau_{r,k}^{(i)} \mid \theta_{r-1}\right)\right],
\end{align}
where $\beta \in (0,1]$ and $w^{(i)}$ is the importance sampling weight, which is defined as: $$w^{(i)}\left(\tau_{r,k}^{(i)} \mid \theta_{r-1}, \theta_{r,k}^{(i)}\right) \triangleq \frac{p^{(i)}\left(\tau_{r,k}^{(i)} \mid \theta_{r-1}\right)}{p^{(i)}\left(\tau_{r,k}^{(i)} \mid \theta_{r,k}^{(i)}\right)}.$$
When $\beta =1,$ Eq.~\eqref{eq:gradient_local} reduces to the stochastic PG direction. When $\beta=0,$ it reduces to the variance-reduced PG direction. Notably, compared to the \textsc{IS-MBPG} algorithm of~\citet{huang2020momentum} for the centralized RL setting, the local updating rule in Algorithm~\ref{algFedRL} differs in that we estimate the PG directions locally, $\theta_{r,k}^{(i)}$, and globally $\theta_{r-1},$ instead of two consecutive local policies. Furthermore, \textsc{FEDSVRPG-M} only requires constant local step-sizes, in contrast to the decreasing step-sizes in~\citet{huang2020momentum}. Moreover, \textsc{FedSVRPG-M} only samples \emph{one trajectory} per iterate, i.e., not does not require very large batch sizes, which is often necessary for  centralized variance-reduced PG methods~\citep{xu2019sample,yuan2020stochastic}. For more discussion on the variance-reduced PG-type algorithms, we refer readers to~\citet{gargiani2022page}.

A notable feature of \textsc{FEDSVRPG-M} is communication efficiency and data locality. To save the communication costs and preserve privacy, all agents upload their local model's difference $\Delta_{r}^{(i)},$ instead of the raw trajectories, to the server only after $K$ local iterations (line 10). Following this step, the server aggregates all the differences to update the global model $\theta_{r+1}$ using the global step-size $\lambda$ and then broadcasts it to all agents. Note that \textsc{FedSVRPG-M} follows the same structure of the vanilla \textsc{FedAVG} and achieves the same communication cost per communication round as \textsc{FedAVG}. 

\textbf{Comparison with prior work.} Note that the algorithms in~\citet{fan2021fault} require the server  to own its own environment (an MDP). They utilized the variance-reduced PG method for updating global models on the server side and applied the stochastic PG method to update the local model \emph{only once} on the agent side. In contrast, our algorithms eliminate the need for the server to own its environment, enhancing its applicability in real-world scenarios. This is crucial as, in numerous cases, the server may function as a third-party entity without access to the environment.

\textbf{Challenges.} Most importantly, our algorithms \emph{accommodate multiple local updates}, a crucial step for reducing the communication costs in FL. Thus, it is important for us to mitigate the common \textit{``client-drift''} problems due to heterogeneity among agents. Notably, even for the standard FL algorithms in the supervised setting, it takes a substantial effort for the FL community to tackle this problem, such as~\textsc{FedProx}~\citep{li2020federated}, \textsc{FedNova}~\citep{wang2020tackling}, \textsc{Scaffold}~\citep{karimireddy2020scaffold} and \textsc{FedLin}~\citep{mitra2021linear}. This challenge is further exacerbated in FRL, where the \emph{non-oblivious} nature of problems makes it uncertain whether the bounded gradient heterogeneity assumption, commonly employed in FL optimization literature, remains applicable. Consequently, achieving a balance between communication cost and convergence rate is challenging. We analyze the performance of \textsc{FEDSVRPG-M} in Section~\ref{sc:main_theorem}.

\begin{algorithm}[tb]
\caption{Description of \textsc{FedSVRPG-M}}
\label{algFedRL}
\begin{algorithmic}
 \STATE {\bfseries{Input:}} initial model $\theta_{-1}=\theta_0,$ gradient estimate $u_0$, local step-size $\eta$, global step-size $\lambda$ and momentum $\beta$. 
\FOR{$r=0,1,\ldots, R-1$}
\STATE $\rhd$ {Agent side}
\FOR {each agent $i \in [N]$}
\STATE Initial local model $\theta_{r,0}^{(i)} =\theta_r$
\FOR {$k=0,1,\ldots, K-1$}
\STATE Sample a trajectory $\tau_{r,k}^{(i)} \sim p^{(i)}\left(\tau \mid \theta_{r,k}^{(i)}\right)$ and compute $u_{r,k}^{(i)}$ using Eq.~\eqref{eq:gradient_local}.
\STATE Update local model $\theta_{r,k+1}^{(i)} = \theta_{r,k}^{(i)} +\eta u_{r,k}^{(i)}$
\ENDFOR
\STATE Send $\Delta_{r}^{(i)} =\theta_{r,K}^{(i)}-\theta_r $  to the server 
\ENDFOR
\STATE $\rhd$ {Server side}
\STATE Aggregate $u_{r+1} = \frac{1}{\eta NK} \sum_{i=1}^N \Delta_{r}^{(i)} $
\STATE Update global model $\theta_{r+1} =\theta_r +\lambda u_{r+1}$
\ENDFOR
\end{algorithmic}
\end{algorithm}

\subsection{\textsc{FEDHAPG-M}}
Recently, \textsc{HAPG}~\citep{shen2019hessian} has been proposed for the centralized RL to reduce the sample complexity from $\mathcal{O}\left(1 / \epsilon^4\right)$ to $\mathcal{O}\left(1 / \epsilon^3\right)$ to obtain the $\epsilon$-FOSP. The main success of \textsc{HAPG} comes from that it utilizes the stochastic approximation of the second-order policy differential. While \textsc{HAPG} uses curvature information, the computation cost of \textsc{HAPG} is still \textit{linear} per iteration with respect to the parameter dimension $d$, as it avoids computing the Hessian explicitly.

We now provide a federated variant of \textsc{HAPG}; Federated Hessian Aided Policy Gradient with Momentum (\textsc{FedHAPG-M}). As discussed in \textsc{FedSVRPG-M}, the usage of momentum in \textsc{FedHAPG-M} primarily serves to offer an ``anchoring" direction that encodes PG estimates from all agents. Consequently, it eliminates the need for bounded environment heterogeneity assumption in existing FRL literature~\citep{jin2022federated, wang2023federated, xie2023fedkl}. Moreover, \textsc{FEDHAPG-M} employs a second-order approximation instead of computing the difference between two consecutive stochastic gradients. As a result, \textsc{FEDHAPG-M} obtains an improved sample complexity akin to that of \textsc{FedSVRPG-M}.

Note that \textsc{FEDHAPG-M} follows the same structure of the vanilla \textsc{FedAVG} and \textsc{FedSVRPG-M}, differing only in the local update procedure. In \textsc{FEDHAPG-M}, we replace the local update direction in \textsc{FedAVG} with a variant of \textsc{HAPG}, see line $7\sim 9$ in Algorithm~\ref{algFedHAPG}. It is worth noting that the uniform sampling step in line 7 guarantees
that $\Lambda_{r,k}^{(i)}$ is an unbiased estimator of $\nabla J(\theta_{r,k}^{(i)}) -\nabla J(\theta_{r-1})$. To estimate the term $\Lambda_{r,k}^{(i)}$, as in \citet{furmston2016approximate, shen2019hessian}, we first assume that the function $J_i(\theta)$ is twice 
differentiable for all $i \in [N]$. Then we compute it as:
\begin{align}\label{eq: Delta}
\Lambda_{r,k}^{(i)}\triangleq & \left\langle\nabla \log p\left(\tau_{r,k}^{(i)} \mid \theta_{r,k}^{(i)}(\alpha)\right),  v_{r,k}^{(i)}\right\rangle g_i\left(\tau_{r,k}^{(i)} \mid \theta_{r,k}^{(i)}(\alpha)\right)\notag\\
&+\nabla \left\langle g_i\left(\tau_{r,k}^{(i)} \mid \theta_{r,k}^{(i)}(\alpha)\right), v_{r,k}^{(i)}\right\rangle
\end{align}
where $v_{r,k}^{(i)} \triangleq \theta_{r,k}^{(i)} -\theta_{r-1}$. The variable $\theta_{r-1}$
represents the last-iterate global policy
    maintained in the server. As mentioned in~\citet{fatkhullin2023stochastic}, the computation of the second term in Eq~\eqref{eq: Delta} can be simplified through via automatic differentiation of the scalar quantity $g\left(\tau_{r,k}^{(i)} \mid \theta_{r,k}^{(i)}(\alpha)\right).$ Thus, the computation cost of \textsc{FedHAPG-M} does not increase and remains at \( \mathcal{O}(Hd)\).
    
\textbf{Discussion.} Same as \textsc{FedSVRPG-M}, \textsc{FedHAPG-M} enjoys the following favorable features: (1) Only sampling one trajectory per local iteration; (2) No need for the server to have its own environment; (3) Multiple local updates. Such features were not simultaneously addressed in ~\citet{fan2021fault, xie2023client}.

\begin{algorithm}[tb]
\caption{Description of \textsc{FedHAPG-M}}
\label{algFedHAPG}
\begin{algorithmic}
\STATE {\bfseries Input:} initial model $\theta_{-1}=\theta_0$ and gradient estimate $u_0$, local step-size $\eta$, global step-size $\lambda$ and momentum $\beta$. 
\FOR {$r=0, \cdots,R-1$} 
\STATE $\rhd$ {Agent side}
\FOR {each agent $i \in [N]$}
\STATE Initial local model $\theta^{(i)}_{r, 0}=\theta_r$ 
\FOR {$k=0, \cdots, K-1$}
\STATE Choose $\alpha$ uniformly at random from $[0,1]$, and compute $\theta_{r,k}^{(i)}(\alpha)=\alpha \theta_{r-1}+(1-\alpha) \theta_{r,k}^{(i)}$
\STATE Sample a trajectory $\tau_{r,k}^{(i)}$ from the density $p^{(i)}\left(\tau \mid \theta_{r,k}^{(i)}(\alpha)\right)$ and compute
$u_{r,k}^{(i)}=\beta w^{(i)}\left(\tau_{r,k}^{(i)} \mid  \theta_{r,k}^{(i)},\theta_{r,k}^{(i)}(\alpha)\right)g_i\left(\tau_{r,k}^{(i)}\mid \theta_{r,k}^{(i)}\right)+\left(1-\beta\right)\left[u_r+\Lambda_{r,k}^{(i)}\right]$, where $\Lambda_{r,k}^{(i)}$ can be computed by using Eq.~\eqref{eq: Delta}
\STATE Update local model $\theta_{r,k+1}^{(i)} = \theta_{r,k}^{(i)} +\eta u_{r,k}^{(i)}$
 \ENDFOR
\STATE Send $\Delta_{r}^{(i)} =\theta_{r,K}^{(i)}-\theta_r $ back to the server
 \ENDFOR
\STATE $\rhd$ {Server side}
\STATE Aggregate $u_{r+1} = \frac{1}{\eta NK} \sum_{i=1}^N \Delta_{r}^{(i)} $
\STATE Update global model $\theta_{r+1} =\theta_r +\lambda u_{r+1}$
 \ENDFOR
\end{algorithmic}
\end{algorithm}

\section{Convergence Analysis}\label{sc:main_theorem}
First, we introduce some standard assumptions.
\begin{assume}\label{assume_policy}
 Let $\pi^{(i)}_{{\theta}}(a \mid s)$ be the policy of the $i$-th agent at state $s$. There exist constants $G, M>0$ such that the log-density of the policy function satisfies
$$
\left\|\nabla_{{\theta}} \log \pi^{(i)}_{{\theta}}(a \mid s)\right\| \leq G, \quad\left\|\nabla_{{\theta}}^2 \log \pi^{(i)}_{{\theta}}(a \mid s)\right\|_2 \leq M,
$$
for all $a \in \mathcal{A}$ and $s \in \mathcal{S}$ and $i \in [N]$.
\end{assume}

\begin{assume}\label{assume_variance}
For each agent $i \in [N]$, the variance of stochastic gradient $g_i(\tau \mid \theta)$ is bounded, i.e., there exists a constant $\sigma>0$, for all policies $\pi_\theta$ such that $\operatorname{Var}(g_i(\tau \mid \theta))=\mathbb{E}\|g_i(\tau \mid \theta)-\nabla J_i(\theta)\|^2 \leq \sigma^2$.
\end{assume}

\begin{assume}\label{assume_IS} For each agent $i \in [N],$ the variance of importance sampling weight $w^{(i)}\left(\tau \mid {\theta}_1, {\theta}_2\right)$ is bounded, i.e., there exists a constant $W>0$ such that 
$
\operatorname{Var}\left(w^{(i)}\left(\tau \mid {\theta}_1, {\theta}_2\right)\right) \leq W$ holds for any ${\theta}_1, {\theta}_2 \in \mathbb{R}^d$ and  $\tau \sim p^{(i)}\left(\cdot \mid {\theta}_2\right)$.
\end{assume}

Assumption~\ref{assume_policy}, \ref{assume_variance} and \ref{assume_IS} are commonly made in the convergence analysis of PG algorithms and their variance-reduced variants~\citep{papini2018stochastic, xu2019sample,shen2019hessian,liu2020improved}. They can be easily verified for Gaussian policies~\citep{cortes2010learning,pirotta2013adaptive, papini2018stochastic}. With these assumptions, we are ready to present the convergence guarantees for our \textsc{Fedsvrpg-m} algorithms.

\begin{theorem} (\textsc{FedSVRPG-M})\label{thm:fedsvrpg_m} 
Under Assumption~\ref{assume_policy}--\ref{assume_IS}, let $u_0=\frac{1}{N B} \sum_{i=1}^N \sum_{b=1}^B g_i\left(\tau_b^{(i)}|\theta_0\right)$ with $B=\left\lceil\frac{K}{R \beta^2}\right\rceil$ and $\left\{\tau^{(i)}_b\right\}_{b=1}^B \stackrel{iid}{\sim} p^{(i)}(\tau | \theta_0)$. There exists a constant local step-size $\eta$, a proper global step-size $\lambda$ and momentum coefficient $\beta$, such that the output of \textsc{FEDSVRPG-M} after $R$ rounds satisfies:
\begin{align}\label{eq:fedsvrpg_theorem}
\frac{1}{R} \sum_{r=0}^{R-1} \mathbb{E}\left[\left\|\nabla J\left(\theta_r\right)\right\|^2\right] \lesssim\left(\frac{\bar{L} \Delta \sigma}{N K R}\right)^{2 / 3}+\frac{\bar{L}\Delta}{R}
\end{align}
where $\Delta \triangleq J\left(\theta^*\right)-J(\theta_0),\ G_0 \triangleq \frac{1}{N} \sum_{i=1}^N\left\|\nabla J_i\left(\theta_0\right)\right\|^2$. 
\end{theorem}
Note that $\bar{L}$ in Theorem~\ref{thm:fedsvrpg_m} is a constant depending on the constants $G, M, W, H,R_{\max}$ and $\frac{1}{(1-\gamma)^2}$. See Appendix for details. The notation $\lesssim$ denotes that inequalities hold up to some numeric number.

\textbf{Comparison with prior work in FRL.}
\textsc{FedSVRPG-M} surpasses all existing results in FRL in convergence, as shown in Table~\ref{tab:1}. Specifically, the results in Theorem 6 from~\citet{jin2022federated} achieve only inexact convergence to a suboptimal solution, depending on the heterogeneity levels among $N$ agents. In contrast, \textsc{FedSVRPG-M} exactly converges to the 
$\epsilon$-FOSP of Problem~\eqref{eq:main_formulation}, with no heterogeneity term observed in Eq.~\eqref{eq:fedsvrpg_theorem}. \citet{fan2021fault} exclusively considered the homogeneous environment. However, their results are limited to the sublinear result. i.e., the stationary point optimality can be scaled by $N^{\frac{2}{3}}$. In contrast, the dominant term $\left(\frac{\bar{L} \Delta \sigma}{N K R}\right)^{2 / 3}$ in the right-hand side of FiEq.~\eqref{eq:fedsvrpg_theorem}  demonstrates that our algorithm provides a $N$-fold linear speedup over the single-agent scenario. Unique to our algorithm is the fact that this speed up is agnostic to the heterogeneity levels, unlike~\citet{woo2023blessing} and \citet{wang2023federated} which obtain a speedup in the no and low heterogeneity regimes respectively.

\textbf{Comparison with prior work in RL.} 
Compared to the centralized RL, i.e., $N=1$, \textsc{FedSVRPG-M} exhibits a convergence rate of $\mathcal{O}\left(1/(K R)^{\frac{2}{3}}\right)$, which aligns with the near-optimal convergence rate in~\citet{fatkhullin2023stochastic}. In contrast, \citet{huang2020momentum}, utilizing diminishing step-sizes, achieves a slower convergence rate of $\mathcal{O}\left(\log(KR)/(K R)^{\frac{2}{3}}\right)$. 

\textbf{Comparison with prior work in FL optimization.}
To appreciate the tightness of our results, we note that our results align with the state-of-the-art convergence rates~\citet{cheng2023momentum,huang2023stochastic} in the FL optimization literature. However, our results are established for a more complex RL setting. In contrast to the supervised learning scenario, where the distribution of $\tau$ is fixed over all iterations, our problem is \emph{non-oblivious}. Furthermore, \text{FEDSVRPG-M} allows for the constant local step-sizes. In contrast, many FL optimization algorithms~\citep{yang2021achieving,khodadadian2022federated} require the decreasing local step-sizes to mitigate heterogeneity among agents. 
\begin{table*}[tb]
        \caption{Impact of environment heterogeneity $\kappa$ and momentum coefficient $\beta$. We evaluate \textsc{FedSVRPG-M} with various $\kappa$ and various momentum coefficient $\beta$ in $\{0.1,0.2,0.5, 0.8\}$. The baseline method is denoted by $\beta =1$. Larger $\kappa$ denotes larger environment heterogeneity. Each setting was run with 16,000 random seeds.}
        \vskip 0.1in
        \vskip 0.15in
\begin{center}
\begin{small}
\begin{sc}
 \begin{tabular}{lccccccc}
        \toprule
        & \multicolumn{6}{c}{Random MDPs} \\
         \cmidrule(r){2-7}
        & \( \kappa = 0 \)& \( \kappa = 0.2 \)& \( \kappa = 0.4 \)& \(\kappa = 0.6 \) &\(\kappa = 0.8 \)&\(\kappa = 1.0 \)\\
        \midrule
        \( \beta = 0.1\)  & \( \mathbf{8.013_{\pm 0.07}} \) & \( \mathbf{7.957_{\pm 0.07}} \) & \( \mathbf{7.968_{\pm 0.06}} \)& \( \mathbf{7.961_{\pm 0.06}}\)&\( \mathbf{7.964_{\pm 0.07}}\) &\( \mathbf{7.981_{\pm 0.06}}\)\\
        \( \beta = 0.2 \) & \( 7.876_{\pm 0.06} \) & \( 7.877_{\pm 0.06} \) & \( 7.851 _{\pm 0.06 }\) & \( 7.837_{\pm 0.06} \)& \( 7.841_{\pm 0.06} \)& \( 7.824_{\pm 0.07} \)\\
        \( \beta = 0.5 \) & \( 7.561_{\pm 0.07} \) & \( 7.208_{\pm 0.06} \) & \( 7.529_{\pm 0.07} \) & \( 7.525_{\pm 0.06} \)& \( 7.536_{\pm 0.07} \)& \( 7.525_{\pm 0.06} \)\\
        \( \beta = 0.8 \) & \( 7.211_{\pm 0.07} \) & \( 7.203_{\pm 0.07} \) & \( 7.201_{\pm 0.06} \)& \( 7.192_{\pm 0.06} \)& \( 7.193_{\pm 0.06} \)& \( 7.184_{\pm 0.06} \) \\
     \midrule  \( \beta=1.0\) & \( 6.965_{\pm 0.07} \) & \( 6.951 _{\pm 0.06} \) & \( 6.955_{\pm 0.06} \)& \( 6.936 _{\pm 0.06} \)& \( 6.940_{\pm 0.06} \)& \( 6.937 _{\pm 0.07} \)\\
        \bottomrule
    \end{tabular}
    \label{tab:comparison}
    \end{sc}
\end{small}
\end{center}
\end{table*}

Now, we analyze the convergence of \textsc{FedHAPG-M}.
\begin{theorem}\label{thm:fedhapg} (\textsc{FEDHAPG-M})
Under Assumption~\ref{assume_policy}--\ref{assume_IS},  choose the same $u_0$ as Theorem~\ref{thm:fedsvrpg_m}.
There exists a constant local step-size $\eta$, a proper global step-size $\lambda$ and momentum coefficient $\beta$, such that the output of \textsc{FEDHAPG-M} after $R$ rounds satisfies
\begin{align}\label{eq:fedhapg_theorem}
\frac{1}{R} \sum_{r=0}^{R-1} \mathbb{E}\left[\left\|\nabla J\left(\theta_r\right)\right\|^2\right] \lesssim\left(\frac{\hat{L} \Delta \sigma}{N K R}\right)^{2 / 3}+\frac{\hat{L}\Delta}{R}
\end{align}
where $\Delta \triangleq J\left(\theta^*\right)-J(\theta_0),\ G_0 \triangleq \frac{1}{N} \sum_{i=1}^N\left\|\nabla J_i\left(\theta_0\right)\right\|^2$
\end{theorem}

From Theorem~\ref{thm:fedhapg}, we remark that \textsc{FEDHAPG-M} enjoys the same worst-case convergence rate, i.e., $\mathcal{O}(1/(NKR)^{2/3})$, as~\textsc{FedSVRPG-M}, except for the differences in the constant $\hat{L}$ and parameter selection. Interested readers are referred to Appendix for details. 

Based on Theorem~\ref{thm:fedsvrpg_m} and~\ref{thm:fedhapg}, we can now translate the convergence results to the total sample complexity of each agent, which is shown in the following corollary.

\begin{corollary}\label{coro:sample_complexity}
Under Assumption~\ref{assume_policy}--\ref{assume_IS}, the sample complexity of \textsc{FedSVRPG-M} and \textsc{FedHAPG-M} is $\mathcal{O}\left(\epsilon^{-\frac{3}{2}}/N\right)$ per agent to find an $\epsilon$-FOSP.
\end{corollary}


\section{Experiments}
We first use tabular environments to verify our theories on the proposed \textsc{FedSVRPG-M} algorithm. It is important to note that  \textsc{FedHAPG-M} algorithm can not be assessed in the tabular setting due to the objective function $J_i(\theta)$ not being twice differentiable. We then evaluate both \textsc{FedSVRPG-M} and \textsc{FedHAPG-M}'s performance on MuJoCo \citep{todorov2012mujoco} with a deep RL extension. The baseline algorithm is the \textsc{PAvg} algorithm~\citep{jin2022federated}.
\begin{figure*}
    \centering
    \includegraphics[width=0.4\textwidth]{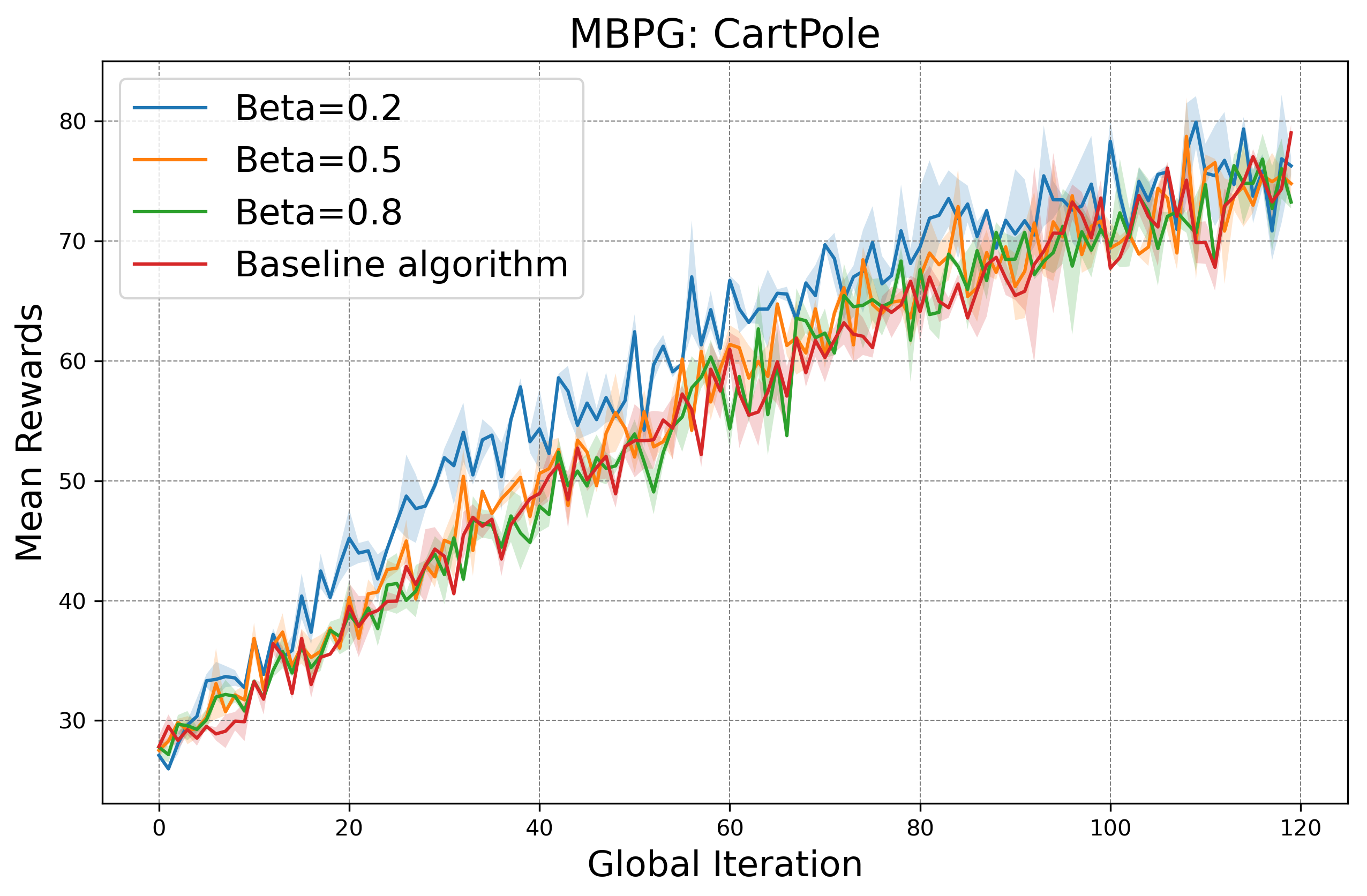}
    \includegraphics[width=0.4\textwidth]{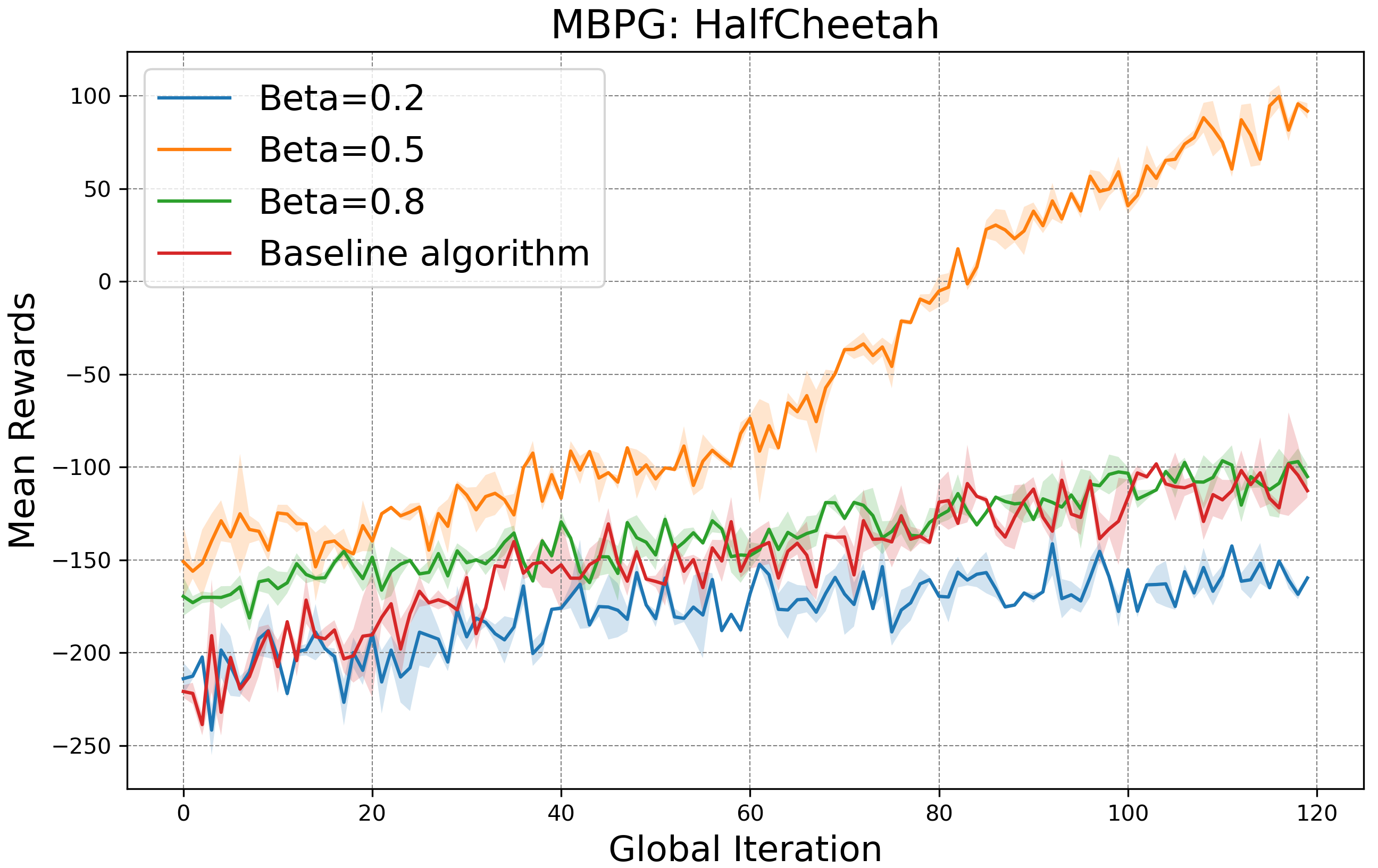}
     \includegraphics[width=0.4\textwidth]{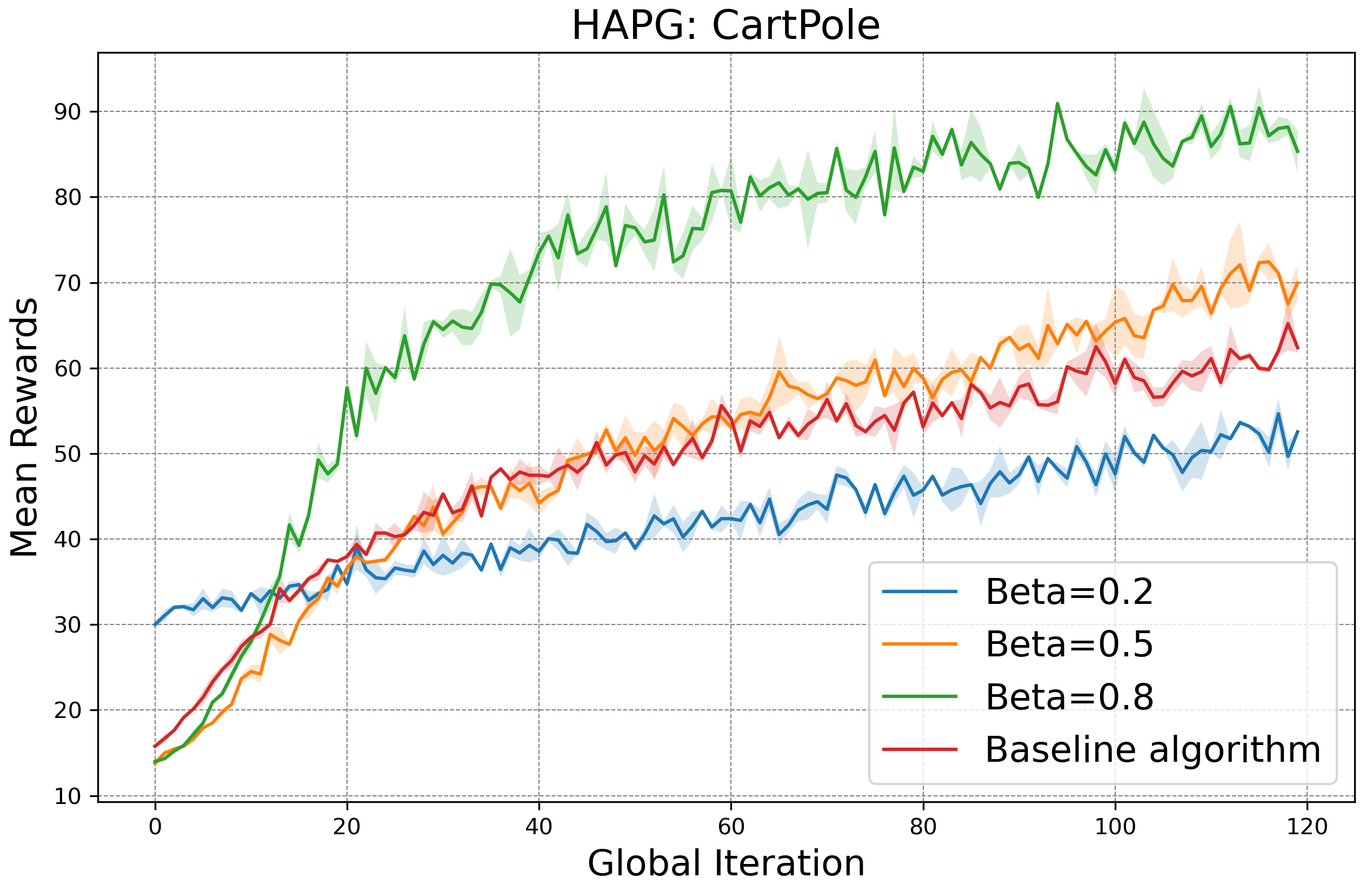}
      \includegraphics[width=0.4\textwidth]{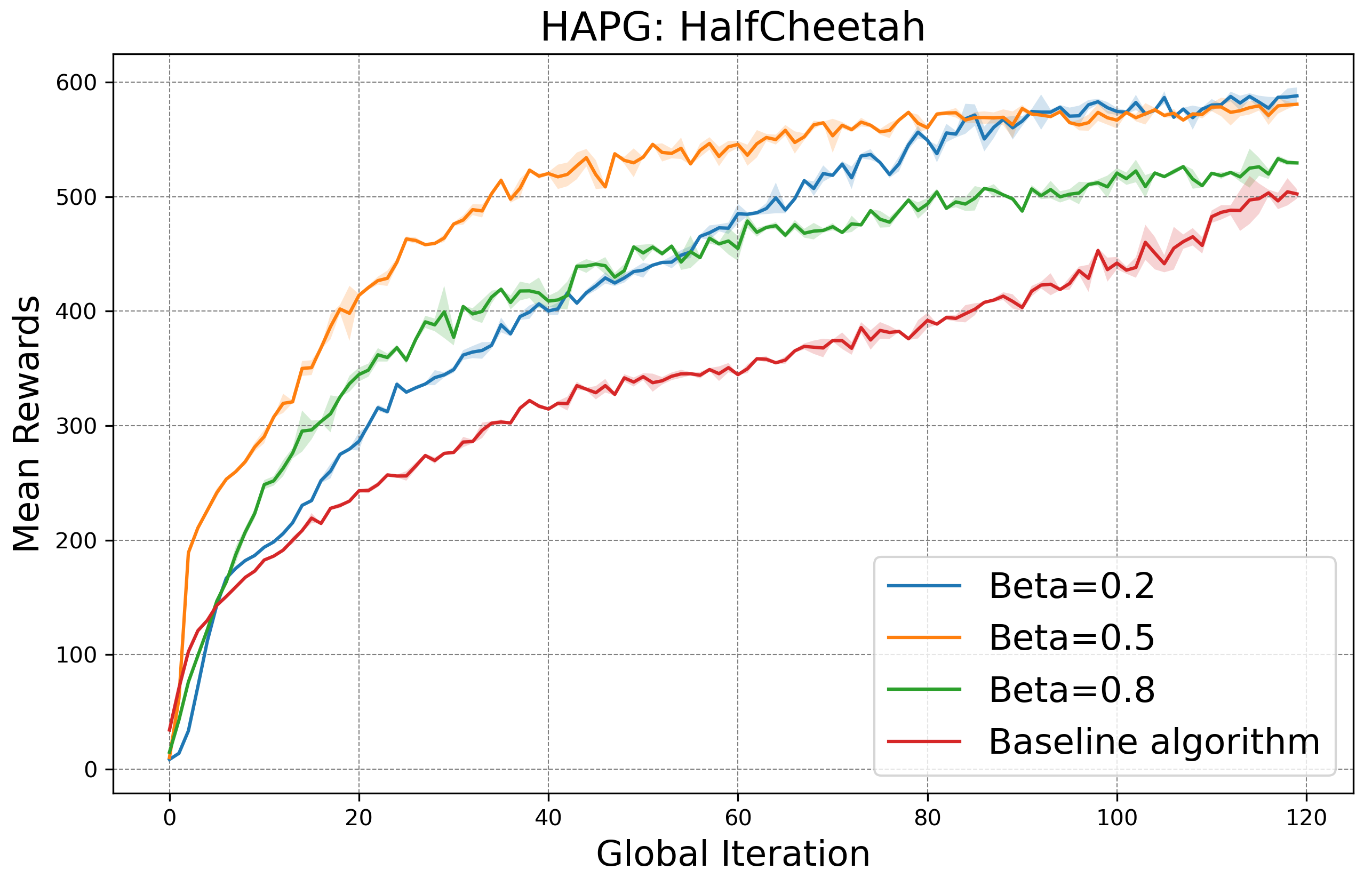}
    \caption{Mean rewards over global iterations for the CartPole and HalfCheetah tasks: (\textbf{Top}): \textsc{FedSVRPG-M}; (\textbf{Bottom}): \textsc{FedHAPG-M}.}
    \label{fig:deep_rl}
    \vspace{-10pt}
\end{figure*}

\textbf{Tabular Case.}
We evaluate the performance of our algorithms in the environment of random MDPs, where both state transitions and reward functions are generated randomly. 
We use the same method as~\citet{jin2022federated} to control the environment heterogeneity. First, we randomly sample a nominal state transition kernel $\mathcal{P}_0$ and then generate the environments $\left\{\mathcal{P}^{(i)} = \kappa \mathcal{P}_{i} + (1-\kappa)\mathcal{P}_0\right\}_{i=1}^N.$ Each entry of the kernels $\{\mathcal{P}_i\}_{i=1}^N$ are uniformly sampled between $0$ and $1$ and then normalized. Then, we can evaluate the impact of environment heterogeneity by varying $\kappa.$ We compare the performance of \textsc{FedSVRPG-M} with the existing baseline algorithm (\textsc{PAVG}). The results are shown in Table~\ref{tab:comparison}. The performance is measured by the average performance function in Eq.~\eqref{eq:main_formulation}. We observe that \textsc{FedSVRPG-M} with $\beta=0.1$ outperforms the baseline algorithm ($\beta=1$). Furthermore, the performance of \textsc{FedSVRPG-M} is agnostic to the environment heterogeneity level $\kappa.$ These trends are expected and consistent with theoretical analysis in Sec.~\ref{sc:main_theorem}.

\textbf{Deep RL Case.} 
We evaluate the performance of our algorithms across two benchmark RL tasks: CartPole and HalfCheetah. While CartPole is a classic control task with discrete actions, HalfCheetah represents a continuous RL task. Both are widely recognized tasks in the MuJoCo simulation environment \citep{todorov2012mujoco}. Comprehensive details of the experimental setups can be found in the appendix. To introduce environment heterogeneity, we change the initial state distribution parameters in both tasks. 
We use Categorical Policy for CartPole, and Gaussian Policy for HalfCheetah. All policies are parameterized by the fully connected neural network which has two hidden layers and a {hyperbolic tangent} activation function.
The hidden layers neural network sizes are 32 for Gaussian policies and 8 for Categorical policies. In Figure \ref{fig:deep_rl}, we show how the mean rewards change over the global iterations for our proposed algorithms and baseline algorithm. 
In both tasks, as the number of iterations increases, all algorithms exhibit a rising trend in mean rewards. There exist a $\beta \neq 1$ that our proposed algorithms outperform the baseline algorithm. In particular, \textsc{FEDSVRPG-M} exhibits optimal performance at $\beta =0.2$ for CartPole and 
$\beta =0.5$ for HalfCheetah.  In contrast, \textsc{FEDHAPG-M} performs optimally with $\beta=0.8$ for CartPole and $\beta=0.5$ for HalfCheetah. \textsc{FEDHAPG-M}, which uses second-order information, shows smaller variance than \textsc{FEDSVRPG-M}, as indicated by the narrower color-shaded regions in the figure. Overall, our algorithms demonstrated superior performance compared to the baseline. See Appendix for more experiments evaluating the linear speedup in the number of agents $N$.

\section{Conclusion}
We introduced \textsc{FEDSVRPG-M} and \textsc{FEDHAPG-M}, overcoming the limitation of bounded environment heterogeneity assumed in prior FRL research. Our results demonstrate the best known convergence for these algorithms and highlight the benefits of collaboration in FRL, even in scenarios with conflicting rewards across different environments. In the future, we plan to focus on algorithms that facilitate downstream fine-tuning or personalization, aiming to discover each MDP's optimal policy through FRL, rather than seeking a universally optimal policy.

\section*{Acknowledgements}

Han Wang and James Anderson are supported by  NSF awards NSF awards ECCS 2144634 and ECCS 2047213. 

Sihong He, Zhili Zhang, and Fei Miao are supported by the National Science Foundation under Grant CNS-2047354 (CAREER).

We thank Xinmeng Huang for many helpful comments
and suggestions.

\section*{Impact Statement}

This paper presents work whose goal is to advance the field of Machine Learning. There are many potential societal consequences of our work, none which we feel must be specifically highlighted here.

\bibliography{reference}
\bibliographystyle{icml2024}

\newpage
\appendix
\onecolumn





\allowdisplaybreaks
\section{Notation}
We denote $\mathcal{F}_0=\emptyset$ and $\mathcal{F}_{r, k}^{(i)}:=\sigma\left(\left\{\theta^{(i)}_{r, j}\right\}_{0 \leq j \leq k} \cup \mathcal{F}_r\right)$ and $\mathcal{F}_{r+1}:=\sigma\left(\cup_i \mathcal{F}^{(i)}_{r, K}\right)$ for all $r \geq 0$ where $\sigma(\cdot)$ indicates the $\sigma$-algebra. Let $\mathbb{E}_r[\cdot]:=\mathbb{E}\left[\cdot \mid \mathcal{F}_r\right]$ be the expectation, conditioned on the filtration $\mathcal{F}_r$, with respect to the random variables $\left\{\tau^{(i)}_{r, k}\right\}_{1 \leq i \leq N, 0 \leq k<K}$ in the $r$-th iteration. Moreover, we use $\mathbb{E}[\cdot]$ to denote the global expectation over all randomness in algorithms.
For all $r \ge 0,$ we define the following notations to simpify the proof:
\begin{align*}
\e_r & :=\mathbb{E}\left[\left\|\nabla J\left(\theta_r\right)-u_{r+1}\right\|^2\right], \\
\mathcal{D}_r & :=\frac{1}{N K} \sum_i \sum_k \mathbb{E}\left[\left\|\theta_{r, k}^{(i)}-\theta_r\right\|\right]^2, \\
c_{r,k}^{(i)} & :=\mathbb{E}\left[\theta_{r, k+1}^{(i)}-\theta_{r, k}^{(i)} \mid \mathcal{F}_{r, k}^{(i)}\right], \\
\mathcal{M}_r & :=\frac{1}{N} \sum_{i=1}^N \mathbb{E}\left[\left\|c_{r, 0}^{(i)}\right\|^2\right].
\end{align*}
Throughout  the appendix, we denote $$\Delta:=J\left(\theta^*\right)-J(\theta_0),\ G_0:=\frac{1}{N} \sum\left\|\nabla J_i\left(\theta_0\right)\right\|^2, \ \theta_{-1}:=\theta_0 \ \text { and } \ \e_{-1}:=\mathbb{E}\left[\left\|\nabla J\left(\theta_0\right)-u_0\right\|^2\right].$$
and $\theta^{*}$ denotes the optimal policy of the optimization problem $(3).$

\section{Useful Lemmas and Inequalities}
We make repeated use throughout the appendix (often without explicitly stating so) of the following inequalities:

\begin{itemize}
 \item  Given any two vectors $x, y \in \mathbb{R}^d$,  for any $a>0$, we have
\begin{equation}\label{eq:youngs}
\|x+y\|^2 \leq(1+a)\|x\|^2+\left(1+\frac{1}{a}\right)\|y\|^2 .
\end{equation}
\item Given any two vectors $x, y \in \mathbb{R}^d$,  for any constant $a>0$, we have
\begin{equation}\label{eq:youngs_inner}
\langle x, y\rangle \le \frac{a}{2}\lVert x\rVert^2 +\frac{1}{2a}\lVert y \rVert^2.
\end{equation}
This inequality goes by the name of Young's inequality.
 \item  Given $m$ vectors $x_1, \ldots, x_m \in \mathbb{R}^d$, the following is a simple application of Jensen's inequality:
\begin{equation}\label{eq:sum_expand}
\left\|\sum_{i=1}^m x_i\right\|^2 \leq m \sum_{i=1}^m\left\|x_i\right\|^2 .
\end{equation}
\end{itemize}

\vspace*{3em}
\begin{prop} (Proposition 5.2 in~\cite{xu2020improved}) \label{prop:lipschitz}
Under Assumption 1, both $J(\theta)$ and $\{J_i(\theta)\}_{i=1}^N$ are $L$-smooth with $L=H R_{\max}\left(M+H G^2\right) /(1-\lambda)$. In addition, for all ${\theta}_1, {\theta}_2 \in \mathbb{R}^d$, we have
$$
\left\|g_i\left(\tau \mid {\theta}_1\right)-g_i\left(\tau \mid {\theta}_2\right)\right\|_2 \leq L_g\left\|{\theta}_1-{\theta}_2\right\|_2
$$
and $\|g_i(\tau \mid {\theta})\|_2 \leq C_g$ for all ${\theta} \in \mathbb{R}^d$ and $i \in [N]$, where $L_g=$ $H MR_{\max} /(1-\lambda), C_g=H G R_{\max} /(1-\lambda)$.
\end{prop}

\vspace*{3em}
\begin{lemma}\label{lem:function_decay}
 If $\lambda L \leq \frac{1}{24}$, the following inequality holds for all $r \geq 0$ :
$$
\mathbb{E}\left[J\left(\theta_{r+1}\right)\right] \ge \mathbb{E}\left[J\left(\theta_r\right)\right]+\frac{11 \lambda}{24} \mathbb{E}\left[\left\|\nabla J\left(\theta_r\right)\right\|^2\right]-\frac{13 \lambda}{24} \e_r .
$$
\end{lemma}
\begin{proof}
Since $J$ is $L$-smooth, we have
\begin{align*}
J\left(\theta_{r+1}\right) & \ge J\left(\theta_r\right)+\left\langle\nabla J\left(\theta_r\right), \theta_{r+1}-\theta_r\right\rangle-\frac{L}{2}\left\|\theta_{r+1}-\theta_r\right\|^2 \\
&=J\left(\theta_r\right)+\lambda\left\|\nabla J\left(\theta_r\right)\right\|^2+\lambda\left\langle\nabla J\left(\theta_r\right), u_{r+1} - \nabla J\left(\theta_r\right)\right\rangle-\frac{L \lambda^2}{2}\left\|u_{r+1}\right\|^2 .
\end{align*}
where we use the fact that $\theta_{r+1} =\theta_r + \eta u_{r+1}$. By using Young's inequality, we have
\begin{align*}
& J\left(\theta_{r+1}\right) \notag\\
\ge & J\left(\theta_r\right)+\frac{\lambda}{2}\left\|\nabla J\left(\theta_r\right)\right\|^2-\frac{\lambda}{2}\left\|\nabla J\left(\theta_r\right)-u_{r+1}\right\|^2 - L \lambda^2\left(\left\|\nabla J\left(\theta_r\right)\right\|^2+\left\|\nabla J\left(\theta_r\right)-u_{r+1}\right\|^2\right) \\
\ge & J\left(\theta_r\right)+ \frac{11 \lambda}{24}\left\|\nabla J\left(\theta_r\right)\right\|^2-\frac{13 \lambda}{24}\left\|\nabla J\left(\theta_r\right)-u_{r+1}\right\|^2,
\end{align*}
where the last inequality holds due to $\lambda L \leq \frac{1}{24}$. Taking the global expectation completes the proof.
\end{proof}

\vspace*{3em}
\begin{lemma} (Lemma 6.1 in~\cite{xu2020improved})
Under Assumptions~\ref{assume_policy} and~\ref{assume_IS}, we have
$$
\operatorname{Var}\left(w^{(i)}(\tau \mid \theta_1, \theta_2) \right) \leq C_w\left\|\theta_1 -\theta_2\right\|^2
$$ holds for any $\theta_1, \theta_2 \in \mathcal{R}^d$ and any $i \in [N],$ where $C_\omega=H\left(2 H G^2+M\right)(W+1)$. 
\end{lemma}

\vspace*{3em}
\section{Federated Stochastic Variance-Reduced Policy Gradient with Momentum}
According to the updating rule of \textsc{FEDSVRPG-M}, we have
$$\E[u_{r+1}] =\frac{1}{NK} \sum_{i,k} \E\left[\nabla J_i(\theta_{r,k}^{(i)})+(1-\beta)\left(u_r -\nabla J_i (\theta_r)\right)\right].$$

\begin{lemma}\label{lem:gradient_bias_SVRG}
If $\lambda \le \sqrt{\frac{16\beta NK}{\widetilde{L_2}^2}},$ we have 
\begin{align*}
\Sigma_{r} \le (1-\frac{8\beta}{9})\Sigma_{r-1} + \frac{\widetilde{L_1}^2}{\beta} \mathcal{D}_r + \frac{3\beta^2 \sigma^2}{NK} +18\lambda^2\frac{\widetilde{L_2}^2}{NK}\E\bigg\|\nabla J(\theta_{r-1}) \bigg\|^2 
\end{align*}
holds for $r\ge 1,$ where $\widetilde{L_1}^2: =L^2+24C_w C_g^2 + 6L_g^2$ and $\widetilde{L_2}^2: =L_g^2 +2C_w C_g^2.$ When $r=0$, we have 
\begin{align*}
\Sigma_{0} &\le (1-\beta)\Sigma_{-1} +\frac{\widetilde{L_1}^2}{\beta} \mathcal{D}_0 + \frac{3\beta^2 \sigma^2}{NK} 
\end{align*}
\end{lemma}
\begin{proof}
\begin{align*} 
&\e_r = \E\left[\left\lVert  u_{r+1} -\nabla J(\theta_r)  \right\rVert^2\right] \\
& = \E \left[ \Big\lVert \frac{1}{NK}\sum_{i,k}\beta g_i\left(\tau_{r,k}^{(i)}\mid \theta_{r,k}^{(i)}\right)+\left(1-\beta\right)\left[u_r+g_i\left(\tau_{r,k}^{(i)}\mid \theta_{r,k}^{(i)}\right)- w^{(i)}\left(\tau_{r,k}^{(i)} \mid \theta_{r-1}, \theta_{r,k}^{(i)}\right) g_i\left(\tau_{r,k}^{(i)} \mid \theta_{r-1}\right)\right]\right.\\
&\left.-\nabla J(\theta_r) \Big\rVert^2\right] \\
& = \E\left[\Big\lVert (1-\beta)(u_r -\nabla J(\theta_{r-1})) +\frac{1}{NK}\sum_{i,k}\left[g_i\left(\tau_{r,k}^{(i)}\mid \theta_{r,k}^{(i)}\right) -\nabla J (\theta_r)\right]\right.\\
& \left.- (1-\beta)\left[\frac{1}{NK}\sum_{i,k} w^{(i)}\left(\tau_{r,k}^{(i)} \mid \theta_{r-1}, \theta_{r,k}^{(i)}\right) g_i\left(\tau_{r,k}^{(i)} \mid \theta_{r-1}\right)-\nabla J(\theta_{r-1})\right] \Big\rVert^2\right] \\
& = (1-\beta)^2 \e_{r-1} + \underbrace{2\E\left[\Big\langle(1-\beta)(u_r -\nabla J(\theta_{r-1})), \frac{1}{NK}\sum_{i,k}\left[\nabla J_i(\theta_{r,k}^{(i)}) -\nabla J (\theta_r)\right]\Big\rangle \right]}_{T_1}\\
& + \underbrace{\E\left\| \frac{1}{NK}\sum_{i,k}\left[g_i\left(\tau_{r,k}^{(i)}\mid \theta_{r,k}^{(i)}\right) -\nabla J (\theta_r)\right]  - (1-\beta)\left[\frac{1}{NK} \sum_{i,k} w^{(i)}\left(\tau_{r,k}^{(i)} \mid \theta_{r-1}, \theta_{r,k}^{(i)}\right) g_i\left(\tau_{r,k}^{(i)} \mid \theta_{r-1}\right)-\nabla J(\theta_{r-1})\right]\right\|^2}_{T_2}
\end{align*}

Using Young's inequality to bound $T_1$, we have 
\begin{align}
T_1 &\le \beta(1-\beta)^2\E\Big\lVert u_r -\nabla J(\theta_{r-1})\Big\rVert^2 + \frac{1}{\beta}\E\Big\lVert \frac{1}{NK}\sum_{i,k}\nabla J_i(\theta_{r,k}^{(i)}) -\nabla J (\theta_r)\Big\rVert^2\notag\\
&\le \beta(1-\beta)^2 \Sigma_{r-1} + \frac{L^2}{\beta}\underbrace{\frac{1}{NK} \sum_{i,k} \E\left\|\theta_{r,k}^{(i)}-\theta_r \right\|^2}_{\mathcal{D}_r}
\end{align}

Further bounding $T_2$, we have
\begin{align}\label{eq:T_2}
T_2 &\le \E\bigg\| \frac{1}{NK}\sum_{i,k}\left(g_i\left(\tau_{r,k}^{(i)}\mid \theta_{r,k}^{(i)}\right) -w^{(i)}\left(\tau_{r,k}^{(i)} \mid \theta_{r}, \theta_{r,k}^{(i)}\right) g_i\left(\tau_{r,k}^{(i)}\mid \theta_{r}\right)\right) \notag\\
&+ \beta \left(\frac{1}{NK}\sum_{i,k} w^{(i)}\left(\tau_{r,k}^{(i)} \mid \theta_{r}, \theta_{r,k}^{(i)}\right) g_i\left(\tau_{r,k}^{(i)}\mid \theta_{r}\right)-\nabla J(\theta_r)\right)\notag\\
& + (1-\beta)\Big (\frac{1}{NK}
\sum_{i,k}\left( w^{(i)}\left(\tau_{r,k}^{(i)} \mid \theta_{r}, \theta_{r,k}^{(i)}\right) 
 g_i(\tau_{r,k}^{(i)}\mid \theta_{r})- w^{(i)}\left(\tau_{r,k}^{(i)} \mid \theta_{r-1}, \theta_{r,k}^{(i)}\right) 
 g_i(\tau_{r,k}^{(i)}\mid \theta_{r-1}) \right) \notag\\
 &-\nabla J(\theta_r) + \nabla J(\theta_{r-1}) \Big)\bigg\|^2 \notag\\
& \le  3\underbrace{\E\bigg\| \frac{1}{NK}\sum_{i,k}\left(g_i\left(\tau_{r,k}^{(i)}\mid \theta_{r,k}^{(i)}\right) -w^{(i)}\left(\tau_{r,k}^{(i)} \mid \theta_{r}, \theta_{r,k}^{(i)}\right) g_i\left(\tau_{r,k}^{(i)}\mid \theta_{r}\right)\right)\bigg\|^2}_{T_{21}} \notag\\
& + 3\frac{\beta^2\sigma^2}{NK} + 3(1-\beta)^2 \underbrace{\E\left[\left\|\frac{1}{NK}
\sum_{i,k}w^{(i)}\left(\tau_{r,k}^{(i)} \mid \theta_{r}, \theta_{r,k}^{(i)}\right) 
 g_i(\tau_{r,k}^{(i)}\mid \theta_{r})- w^{(i)}\left(\tau_{r,k}^{(i)} \mid \theta_{r-1}, \theta_{r,k}^{(i)}\right) 
 g_i(\tau_{r,k}^{(i)}\mid \theta_{r-1})  \right\|^2 \right]}_{T_{22}} 
\end{align}
where we use the Young's inequality in the last equality and the fact that $\E[\| X -\E[X]\|^2]\le \E[\|X\|^2]$ holds for any random variable $X$.

To precede, we continue to bound $T_{21}$ and have that
\begin{align}\label{eq:T_21}
T_{21} &=\E\bigg\| \frac{1}{NK}\sum_{i,k}\left(g_i\left(\tau_{r,k}^{(i)}\mid \theta_{r,k}^{(i)}\right) -w^{(i)}\left(\tau_{r,k}^{(i)} \mid \theta_{r}, \theta_{r,k}^{(i)}\right) g_i\left(\tau_{r,k}^{(i)}\mid \theta_{r}\right)\right)\bigg\|^2 \notag\\
& \le 2\E\bigg\| \frac{1}{NK}\sum_{i,k}\left(1 -w^{(i)}(\tau_{r,k}^{(i)} \mid \theta_{r}, \theta_{r,k}^{(i)})\right) g_i(\tau_{r,k}^{(i)}\mid \theta_r)\bigg\|^2 \notag\\
& + 2\E\bigg\| \frac{1}{NK}\sum_{i,k}\left[g_i\left(\tau_{r,k}^{(i)}\mid \theta_{r,k}^{(i)}\right)-g_i(\tau_{r,k}^{(i)}\mid \theta_r)\right]\bigg\|^2 \notag\\
& \le \frac{2C_w C_g^2}{NK}\sum_{i,k}  \E\bigg\|\theta_{r,k}^{(i)} -\theta_r\bigg\|^2 + 2\frac{L_g^2}{NK}\sum_{i,k}  \E\bigg\|\theta_{r,k}^{(i)} -\theta_r\bigg\|^2 \notag\\
& = (2C_w C_g^2+2L_g^2) \mathcal{D}_r
\end{align}
where we use the fact that $\|g^{(i)}(\tau \mid \theta)\|_2 \leq C_g \text { for all } \theta \in \mathbb{R}^d$ and $i \in [N]$.

To bound $T_{22},$ we have
\begin{align}\label{eq:T_22}
T_{22}&=\E\left[\left\|\frac{1}{NK}
\sum_{i,k}w^{(i)}\left(\tau_{r,k}^{(i)} \mid \theta_{r}, \theta_{r,k}^{(i)}\right) 
 g_i(\tau_{r,k}^{(i)}\mid \theta_{r})- w^{(i)}\left(\tau_{r,k}^{(i)} \mid \theta_{r-1}, \theta_{r,k}^{(i)}\right) 
 g_i(\tau_{r,k}^{(i)}\mid \theta_{r-1})  \right\|^2 \right] \notag\\
 & \le 3\E\left[\left\|\frac{1}{NK}
\sum_{i,k}\left[w^{(i)}\left(\tau_{r,k}^{(i)} \mid \theta_{r}, \theta_{r,k}^{(i)}\right)-1\right] 
 g_i(\tau_{r,k}^{(i)}\mid \theta_{r})\right\|^2 \right] \notag\\
 & + 3\frac{1}{N^2K^2}\sum_{i,k}\E\left\|g_i(\tau_{r,k}^{(i)}\mid \theta_{r})-g_i(\tau_{r,k}^{(i)}\mid \theta_{r-1})\right\|^2 + 3\E\left[\left\|\frac{1}{NK}
\sum_{i,k}\left[w^{(i)}\left(\tau_{r,k}^{(i)} \mid \theta_{r-1}, \theta_{r,k}^{(i)}\right)-1\right] 
 g_i(\tau_{r,k}^{(i)}\mid \theta_{r-1})\right\|^2 \right] \notag\\
 &\le 3C_g^2 C_w\frac{1}{N^2K^2}\sum_{i,k}\E\bigg\|\theta_{r,k}^{(i)} -\theta_r\bigg\|^2 + 3\frac{L_g^2}{NK}\E\bigg\|\theta_{r-1} -\theta_r\bigg\|^2 + 3C_g^2 C_w\frac{1}{N^2K^2}\sum_{i,k}\E\bigg\|\theta_{r,k}^{(i)} -\theta_{r-1}\bigg\|^2 \notag\\
 & \le 6C_g^2 C_w \frac{1}{NK} \mathcal{D}_r + \frac{3L_g^2 +6C_w C_g^2}{NK}\E\bigg\|\theta_{r-1} -\theta_r\bigg\|^2
\end{align}

Combining the upper bound of $T_{21}$ and $T_{22}$ (i.e., $\eqref{eq:T_21}$ and $\eqref{eq:T_22}$) into $T_2$ in Eq.~\eqref{eq:T_2}, we have
\begin{align}
T_2 \le (24C_w C_g^2 + 6L_g^2)\mathcal{D}_r + \frac{3\beta^2 \sigma^2}{NK} +9(1-\beta)^2 \frac{L_g^2 +2C_w C_g^2}{NK}\E\bigg\|\theta_{r-1} -\theta_r\bigg\|^2
\end{align}
Therefore, for $r\ge 1,$ we have
\begin{align}
\Sigma_{r} &\le (1-\beta)\Sigma_{r-1} +\frac{L^2+24C_w C_g^2 + 6L_g^2}{\beta} \mathcal{D}_r + \frac{3\beta^2 \sigma^2}{NK} +9(1-\beta)^2 \frac{L_g^2 +2C_w C_g^2}{NK}\E\bigg\|\theta_{r-1} -\theta_r\bigg\|^2 \label{eq:bound_0_step}\\
& \le (1-\beta)\Sigma_{r-1} +\frac{L^2+24C_w C_g^2 + 6L_g^2}{\beta} \mathcal{D}_r + \frac{3\beta^2 \sigma^2}{NK}\notag\\
&+18\lambda^2\frac{L_g^2 +2C_w C_g^2}{NK}\E\bigg\|\nabla J(\theta_{r-1}) \bigg\|^2 + 18\lambda^2\frac{L_g^2 +2C_w C_g^2}{NK}\E\bigg\|\nabla J(\theta_{r-1})-u_r \bigg\|^2 \notag\\
& = \left(1-\beta+ 18\lambda^2\frac{L_g^2 +2C_w C_g^2}{NK}\right)\Sigma_{r-1}+18\lambda^2\frac{L_g^2 +2C_w C_g^2}{NK}\E\bigg\|\nabla J(\theta_{r-1}) \bigg\|^2  \notag\\
& + \frac{L^2+24C_w C_g^2 + 6L_g^2}{\beta} \mathcal{D}_r + \frac{3\beta^2 \sigma^2}{NK}
\end{align}
By choosing $\lambda$ such that $18\lambda^2\frac{L_g^2 +2C_w C_g^2}{NK} \le\frac{8\beta}{9}$, which holds when $\lambda \le \sqrt{\frac{16\beta NK}{L_g^2 +2C_w C_g^2}},$ we have
\begin{align}
\Sigma_{r} \le (1-\frac{8\beta}{9})\Sigma_{r-1} + \frac{L^2+24C_w C_g^2 + 6L_g^2}{\beta} \mathcal{D}_r + \frac{3\beta^2 \sigma^2}{NK} +18\lambda^2\frac{L_g^2 +2C_w C_g^2}{NK}\E\bigg\|\nabla J(\theta_{r-1}) \bigg\|^2 
\end{align}
holds for $r> 0$. When $r=0$, we have that 
\begin{align}
\Sigma_{0} &\le (1-\beta)\Sigma_{-1} +\frac{L^2+24C_w C_g^2 + 6L_g^2}{\beta} \mathcal{D}_0 + \frac{3\beta^2 \sigma^2}{NK} 
\end{align}
which can be derived from Eq.\eqref{eq:bound_0_step}.
\end{proof}

\vspace*{3em}
\begin{lemma}\label{lem:drift_term_svrg}
(Bounding drift-term) If the local step-size satisfies $\eta \le \min\{\frac{L}{32e^2\widetilde{L_3}^2 K},\frac{1}{KL}\}$, the drift-term can be upper bounded as: 
\begin{align*}
\mathcal{D}_r \le 4e K^2\mathcal{M}_r + (16\eta^4 K^4L^2 + 8\eta^2 K) \left(\beta^2\sigma^2 + 2 \widetilde{L_3}^2\E\Big\|\theta_{r-1} -\theta_{r}\Big\|^2 \right)
\end{align*}
where $\widetilde{L_3}^2:=2C_w C_g^2+2L_g^2.$
\end{lemma}
\begin{proof}
Define $c_{r,k}^{(i)}:= -\eta \left(\nabla J_i(\theta_{r,k}^{(i)}) +(1-\beta)(u_r -\nabla J_{i}(\theta_{r-1}))\right)$. For any $1\le j \le k-1 \le K-2,$ we have:
\begin{align}
\E\left\|c_{r,j}^{(i)} -c_{r,j-1}^{(i)} \right\|^2 &\le \eta^2 L^2 \E\left\|\theta_{r,j}^{(i)} -\theta_{r,j-1}^{(i)} \right\|^2\notag\\
&= \eta^2 L^2\left(\E\left\|c_{r,j-1}^{(i)}\right\|^2 + \mathbb{E}\left[\operatorname{Var}\left[\theta^{(i)}_{r, j}-\theta^{(i)}_{r, j-1} \mid \mathcal{F}^{(i)}_{r, j-1}\right]\right]\right).
\end{align}
where we use the bias-variance decomposition in the last inequality.
\begin{align}
&\mathbb{E}\left[\operatorname{Var}\left[\theta^{(i)}_{r, j}-\theta^{(i)}_{r, j-1} \mid \mathcal{F}^{(i)}_{r, j-1}\right]\right] \notag\\
&=\eta^2\E\bigg\| g_i\left(\tau_{r,j-1}^{(i)}\mid \theta_{r,j-1}^{(i)}\right) -\nabla J_i(\theta_{r,j-1}^{(i)})\notag\\
&-\left(1-\beta\right)\left[ w^{(i)}\left(\tau_{r,j-1}^{(i)} \mid \theta_{r-1}, \theta_{r,j-1}^{(i)}\right) g_i\left(\tau_{r,j-1}^{(i)} \mid \theta_{r-1}\right)-\nabla J_i (\theta_{r-1})\right]\bigg\|^2\notag\\
&=\eta^2\E\bigg\| \beta \left[g_i\left(\tau_{r,j-1}^{(i)}\mid \theta_{r,j-1}^{(i)}\right) -\nabla J_i(\theta_{r,j-1}^{(i)})\right]\notag\\
&+\left(1-\beta\right)\left[g_i\left(\tau_{r,j-1}^{(i)}\mid \theta_{r,j-1}^{(i)}\right)- w^{(i)}\left(\tau_{r,j-1}^{(i)} \mid \theta_{r-1}, \theta_{r,j-1}^{(i)}\right) g_i\left(\tau_{r,j-1}^{(i)} \mid \theta_{r-1}\right) \right.\notag\\
&\quad \quad \left.
-\left(\nabla J_i(\theta_{r,j-1}^{(i)})
-\nabla J_i(\theta_{r-1})\right)\right]\bigg\|^2 \notag\\
& \le 2\eta^2 \beta^2\sigma^2 \notag\\
& + 2 \eta^2 (1-\beta)^2 \underbrace{\E\bigg\|g_i\left(\tau_{r,j-1}^{(i)}\mid \theta_{r,j-1}^{(i)}\right)- w^{(i)}\left(\tau_{r,j-1}^{(i)} \mid \theta_{r-1}, \theta_{r,j-1}^{(i)}\right) g_i\left(\tau_{r,j-1}^{(i)} \mid \theta_{r-1}\right) \bigg\|^2}_{T_3} \label{eq:drift_variance}
\end{align}
where Eq.\eqref{eq:drift_variance} holds due to the Young's inequality and the fact that $\E[\| X -\E[X]\|^2]\le \E[\|X\|^2].$

To precede, we bound $T_3$ as
\begin{align}\label{eq:T_3}
T_3&=\E\bigg\|g_i\left(\tau_{r,j-1}^{(i)}\mid \theta_{r,j-1}^{(i)}\right)- w^{(i)}\left(\tau_{r,j-1}^{(i)} \mid \theta_{r-1}, \theta_{r,j-1}^{(i)}\right) g_i\left(\tau_{r,j-1}^{(i)} \mid \theta_{r-1}\right) \bigg\|^2 \notag\\
& \le  2\E\bigg\| \left(1 -w^{(i)}(\tau_{r,j-1}^{(i)} \mid \theta_{r}, \theta_{r,j-1}^{(i)})\right) g_i(\tau_{r,j-1}^{(i)}\mid \theta_r)\bigg\|^2 \notag\\
& + 2\E\bigg\| g_i\left(\tau_{r,j-1}^{(i)}\mid \theta_{r,j-1}^{(i)}\right)-g_i(\tau_{r,j-1}^{(i)}\mid \theta_{r-1})\bigg\|^2 \notag\\
& \le 2C_w C_g^2  \E\bigg\|\theta_{r,j-1}^{(i)} -\theta_{r-1}\bigg\|^2 + 2L_g^2 \E\bigg\|\theta_{r,j-1}^{(i)} -\theta_{r-1}\bigg\|^2 \notag\\
& = (2C_w C_g^2+2L_g^2) \E\bigg\|\theta_{r,j-1}^{(i)} -\theta_{r-1}\bigg\|^2
\end{align}
where we use the fact that $\|g^{(i)}(\tau \mid \theta)\|_2 \leq C_g \text { for all } \theta \in \mathbb{R}^d$ and $i \in [N]$.

With the upper bound of $T_3$ and $\widetilde{L_3}^2:=2C_w C_g^2+2L_g^2,$ we have
\begin{align}
\E\left\|c_{r,j}^{(i)} -c_{r,j-1}^{(i)} \right\|^2 & \le \eta^2 L^2\bigg(\E\left\|c_{r,j-1}^{(i)}\right\|^2 +2\eta^2\beta^2\sigma^2 + 2\eta^2(1-\beta)^2 \widetilde{L_3}^2\E\Big\|\theta_{r,j-1}^{(i)} -\theta_{r-1}\Big\|^2 \bigg)\notag\\
& \le \eta^2 L^2\bigg(\E\left\|c_{r,j-1}^{(i)}\right\|^2 +2\eta^2\beta^2\sigma^2 + 4\eta^2 \widetilde{L_3}^2\E\Big\|\theta_{r,j-1}^{(i)} -\theta_{r}\Big\|^2+ 4\eta^2 \widetilde{L_3}^2\E\Big\|\theta_{r-1} -\theta_{r}\Big\|^2\bigg).
\end{align}
Then we have 
\begin{align}
&\E\left\|c_{r,j}^{(i)} \right\|^2 \le (1+\frac{1}{q})\E\left\|c_{r,j-1}^{(i)} \right\|^2 + (1+q)\E\left\|c_{r,j}^{(i)} -c_{r,j-1}^{(i)} \right\|^2 \notag\\
& \le (1+\frac{2}{q})\E\left\|c_{r,j-1}^{(i)} \right\|^2 + (1+q)\eta^2 L^2 \left( 2\eta^2\beta^2\sigma^2 +4\eta^2 \widetilde{L_3}^2\E\Big\|\theta_{r,j-1}^{(i)} -\theta_{r}\Big\|^2+4\eta^2 \widetilde{L_3}^2\E\Big\|\theta_{r-1} -\theta_{r}\Big\|^2 \right) 
\end{align}
where we use the fact that $\eta L \le \frac{1}{K}\le\frac{1}{q+1}$ and let $q=k-1$. By unrolling this recurrence, we have
\begin{align}
\E\left\|c_{r,j}^{(i)} \right\|^2 &\le (1+\frac{2}{k-1})^{j}\E\left\|c_{r,0}^{(i)} \right\|^2  + k\eta^2 L^2\sum_{i=0}^{j-1}(2\eta^2\beta^2\sigma^2 +4\eta^2 \widetilde{L_3}^2\E\Big\|\theta_{r-1} -\theta_{r}\Big\|^2)\Pi_{j^{\prime}=i+1}^{j-1}(1+\frac{2}{k-1}) \notag\\
& + k\eta^2 L^2\sum_{s=0}^{j-1}(4\eta^2 \widetilde{L_3}^2\E\Big\|\theta_{r,s}^{(i)} -\theta_{r}\Big\|^2)\Pi_{j^{\prime}=s+1}^{j-1}(1+\frac{2}{k-1}) \notag\\
& \le (1+\frac{2}{k-1})^{k-1}\E\left\|c_{r,0}^{(i)} \right\|^2  + k\eta^2 L^2\sum_{i=0}^{k-1}(2\eta^2\beta^2\sigma^2 +4\eta^2 \widetilde{L_3}^2\E\Big\|\theta_{r-1} -\theta_{r}\Big\|^2)(1+\frac{2}{k-1})^{k-1} \notag\\
& + k\eta^2 L^2\sum_{j^{\prime}=0}^{j-1}(4\eta^2 \widetilde{L_3}^2\E\Big\|\theta_{r,j^{\prime}}^{(i)} -\theta_{r}\Big\|^2)(1+\frac{2}{k-1})^{k-1}
\end{align}
Based on the inequality $(1+\frac{2}{K-1}^{k-1})\le e^2 \le 8$, we have 
\begin{align}\label{eq:upper_v}
\E\left\|c_{r,j}^{(i)} \right\|^2 &\le e^2\E\left\|c_{r,0}^{(i)} \right\|^2 + 8k^2\eta^4L^2 \left(2\beta^2\sigma^2 + 4 \widetilde{L_3}^2\E\Big\|\theta_{r-1} -\theta_{r}\Big\|^2 \right) + 4e^2 k\eta^4 L^2 \widetilde{L_3}^2\sum_{j^{\prime}=0}^{j-1}\E\Big\|\theta_{r,j^{\prime}}^{(i)} -\theta_{r}\Big\|^2
\end{align}
By Lemma A.3, we have
\begin{align}\label{eq:drift_deviation}
\E\left\|\theta_{r,k}^{(i)} -\theta_r \right\|^2 &\le 2 \E\left\| \sum_{j=0}^{k-1} c_{r,j}^{(i)}\right\|^2 + 
2 \sum_{j=0}^{k-1} \mathbb{E}\left[\operatorname{Var}\left[\theta_{r, j+1}^{(i)}-\theta_{r, j}^{(i)} \mid \mathcal{F}_{r, j}^{(i)}\right]\right] \notag\\
&\stackrel{(a)}{\leq} 2 k \sum_{j=0}^{k-1} \mathbb{E}\left\|  c_{r,j}^{(i)}\right\|^2+2 \sum_{j=0}^{k-1}\left(2 \beta^2 \eta^2 \sigma^2+4\eta^2 \widetilde{L_3}^2\E\Big\|\theta_{r,j}^{(i)} -\theta_{r}\Big\|^2+4\eta^2 \widetilde{L_3}^2\E\Big\|\theta_{r-1} -\theta_{r}\Big\|^2\right)
\end{align}
where $(a)$ is due to Eq.\eqref{eq:drift_variance} and Eq.\eqref{eq:T_3}. Plugging Eq.\eqref{eq:upper_v} into Eq.\eqref{eq:drift_deviation}, we have
\begin{align}
&\E\left\|\theta_{r,k}^{(i)} -\theta_r \right\|^2 \le\notag\\
& 2 k \sum_{j=0}^{k-1}\bigg\{e^2 \E\left\|c_{r,0}^{(i)} \right\|^2 + 8k^2\eta^4L^2 \left(2\beta^2\sigma^2 + 4 \widetilde{L_3}^2\E\Big\|\theta_{r-1} -\theta_{r}\Big\|^2 \right) + 4e^2 k\eta^4 L^2 \widetilde{L_3}^2\sum_{j^{\prime}=0}^{j-1}\E\Big\|\theta_{r,j^{\prime}}^{(i)} -\theta_{r}\Big\|^2\bigg\}\notag\\
&+2 \sum_{j=0}^{k-1}\left(2 \beta^2 \eta^2 \sigma^2+4\eta^2 \widetilde{L_3}^2\E\Big\|\theta_{r,j}^{(i)} -\theta_{r}\Big\|^2+4\eta^2 \widetilde{L_3}^2\E\Big\|\theta_{r-1} -\theta_{r}\Big\|^2\right)
\end{align}
Summing up the above equation over $k=0,\cdots, K-1,$ we have
\begin{align}
\sum_{k=0}^{K-1}\E\left\|\theta_{r,k}^{(i)} -\theta_r \right\|^2 &\le \sum_{k=0}^{K-1} \left\{2k^2 e^2 \E\left\|c_{r,0}^{(i)} \right\|^2 + 16k^4\eta^4L^2 \left(2\beta^2\sigma^2 + 4 \widetilde{L_3}^2\E\Big\|\theta_{r-1} -\theta_{r}\Big\|^2 \right)\right\} \notag\\
& +\sum_{k=0}^{K-1}8e^2 k^2 \eta^4 L^2 \widetilde{L_3}^2\sum_{j=0}^{k-1}\sum_{j^{\prime}=0}^{j-1}\E\Big\|\theta_{r,j^{\prime}}^{(i)} -\theta_{r}\Big\|^2 \notag\\
& +\sum_{k=0}^{K-1} \left(4k \beta^2 \eta^2 \sigma^2 + 8k\eta^2 \widetilde{L_3}^2\E\Big\|\theta_{r-1} -\theta_{r}\Big\|^2+ 8\eta^2 \widetilde{L_3}^2\sum_{j=0}^{k-1}\E\Big\|\theta_{r,j}^{(i)} -\theta_{r}\Big\|^2 \right)\notag\\
&\le 2e K^3 \E\left\|c_{r,0}^{(i)} \right\|^2 + (8\eta^4 K^5L^2 + 4\eta^2 K^2) \left(\beta^2\sigma^2 + 2 \widetilde{L_3}^2\E\Big\|\theta_{r-1} -\theta_{r}\Big\|^2 \right) \notag\\
& +K^2\sum_{k=0}^{K-1}8e^2  \eta^4 L^2 \widetilde{L_3}^2\sum_{j=0}^{K-1}\sum_{j^{\prime}=0}^{K-1}\E\Big\|\theta_{r,j^{\prime}}^{(i)} -\theta_{r}\Big\|^2 +\sum_{k=0}^{K-1}8\eta^2 \widetilde{L_3}^2\sum_{j=0}^{K-1}\E\Big\|\theta_{r,j}^{(i)} -\theta_{r}\Big\|^2 \notag\\
& =2e K^3 \E\left\|c_{r,0}^{(i)} \right\|^2 + (8\eta^4 K^5L^2 + 4\eta^2 K^2) \left(\beta^2\sigma^2 + 2 \widetilde{L_3}^2\E\Big\|\theta_{r-1} -\theta_{r}\Big\|^2 \right) \notag\\
& +(8e^2\eta^4 K^4L^2 \widetilde{L_3}^2+8\eta^2 \widetilde{L_3}^2K)\sum_{j=0}^{K-1}\E\Big\|\theta_{r,j}^{(i)} -\theta_{r}\Big\|^2
\end{align}
With the choice of step-size $\eta$ satisfying $8e^2\eta^4 K^4L^2 \widetilde{L_3}^2+8\eta^2 \widetilde{L_3}^2K \le \frac{1}{2},$ after some rearrangement, we have
\begin{align*}
\frac{1}{2K}\sum_{k=0}^{K-1}\E\left\|\theta_{r,k}^{(i)} -\theta_r \right\|^2 \le 2e K^2 \E\left\|c_{r,0}^{(i)} \right\|^2 + (8\eta^4 K^4L^2 + 4\eta^2 K) \left(\beta^2\sigma^2 + 2 \widetilde{L_3}^2\E\Big\|\theta_{r-1} -\theta_{r}\Big\|^2 \right)
\end{align*}
In summary, we can bound the drift-term as 
\begin{align*}
\mathcal{D}_r \le 4e K^2\underbrace{\frac{1}{N}\sum_{i=1}^N\E\left\|c_{r,0}^{(i)} \right\|^2}_{\mathcal{M}_r} + (16\eta^4 K^4L^2 + 8\eta^2 K) \left(\beta^2\sigma^2 + 2 \widetilde{L_3}^2\E\Big\|\theta_{r-1} -\theta_{r}\Big\|^2 \right)
\end{align*}
\end{proof}

\vspace*{3em}
\begin{lemma}\label{lem:high_order_svrg}
If $\lambda L \le \frac{1}{24}$ and $\eta^2 \left[\frac{289}{72}(1-\beta)^2+8 e(\lambda \beta L R)^2\right] \leq \frac{\beta^2}{288eK^2 \widetilde{L_1}^2},$ we have
\begin{align}
\sum_{r=0}^{R-1}\mathcal{M}_r =\frac{1}{N} \sum_{r=0}^{R-1}\sum_{i=1}^N\E\left\|c_{r,0}^{(i)} \right\|^2\le \frac{\beta^2}{288 e K^2 \widetilde{L_1}^2} \sum_{r=-1}^{R-2}\left(\e_r+\mathbb{E}\left[\left\|\nabla J\left(\theta_r\right)\right\|^2\right]\right)+4 \eta^2 \beta^2 e R G_0 .
\end{align}
where $G_0: =\frac{1}{N} \sum_{i=1}^N \mathbb{E}\left[\left\|\nabla J_i\left(\theta_0\right)\right\|^2\right]$ and $\widetilde{L_1}^2$ is defined in Lemma~\ref{lem:gradient_bias_SVRG}.
\end{lemma}
\begin{proof}
Recall that $c_{r,0}^{(i)}:= -\eta \left(\nabla J_i(\theta_{r}) +(1-\beta)(u_r -\nabla J_{i}(\theta_{r-1}))\right).$ Then, it is straightforward to have
\begin{align}\label{eq:local_drift_high}
\left\|c^{(i)}_{r, 0}\right\|^2 & \leq 2 \eta^2\left((1-\beta)^2\left\|u_r\right\|^2+\left\|\nabla J_i\left(\theta_r\right)-(1-\beta) \nabla J_i(\theta_{r-1})\right\|^2\right) \notag\\
& \le  2 \eta^2(1-\beta)^2\left\|u_r\right\|^2+4\eta^2(1-\beta)^2\left\|\nabla J_i\left(\theta_r\right)-\nabla J_i(\theta_{r-1})\right\|^2+4\eta^2 \beta^2\left\|\nabla J_i(\theta_{r})\right\|^2
\notag\\
& \leq 2 \eta^2(1-\beta)^2\left(1+2(\lambda L)^2\right)\left\|u_r\right\|^2+4 \eta^2 \beta^2\left\|\nabla J_i\left(\theta_r\right)\right\|^2 \notag\\
&\stackrel{(a)}{\leq} \frac{289}{144} \eta^2(1-\beta)^2\left\|u_r\right\|^2+4 \eta^2 \beta^2\left\|\nabla J_i\left(\theta_r\right)\right\|^2 .
\end{align}
where (a) is due to the choice of $\lambda$ such that $\lambda L \le \frac{1}{24}$. 

Using the Young's inequality, we have that for any $\zeta >0$, 
\begin{align*}
\mathbb{E}\left[\left\|\nabla J_i\left(\theta_r\right)\right\|^2\right] & \leq(1+\zeta) \mathbb{E}\left[\left\|\nabla J_i\left(\theta_{r-1}\right)\right\|^2\right]+\left(1+\frac{1}{\zeta}\right)  \mathbb{E}\left\|\nabla J_i\left(\theta_r\right)-\nabla J_i\left(\theta_{r-1}\right)\right\|^2 \notag\\
& \leq(1+\zeta) \mathbb{E}\left[\left\|\nabla J_i\left(\theta_{r-1}\right)\right\|^2\right]+\left(1+\frac{1}{\zeta}\right) L^2\E\left\|\theta_r -\theta_{r-1}\right\|^2 \notag\\
&\leq(1+\zeta) \mathbb{E}\left[\left\|\nabla J_i\left(\theta_{r-1}\right)\right\|^2\right]+2\left(1+\frac{1}{\zeta}\right) (\lambda L)^2\left(\E\left\|u_r -\nabla J(\theta_{r-1})\right\|^2 +\E\left\|\nabla J(\theta_{r}) \right\|^2\right) \notag\\
&=(1+\zeta) \mathbb{E}\left[\left\|\nabla J_i\left(\theta_{r-1}\right)\right\|^2\right]+2\left(1+\frac{1}{\zeta}\right) (\lambda L)^2\left(\e_{r-1}+\E\left\|\nabla J(\theta_{r}) \right\|^2\right) 
\end{align*}
By unrolling the recursive bound, we have
\begin{align*}
\mathbb{E}\left[\left\|\nabla J_i\left(\theta_r\right)\right\|^2\right]  \leq(1+\zeta)^r \mathbb{E}\left[\left\|\nabla J_i\left(\theta_0\right)\right\|^2\right]+\frac{2}{\zeta}(\lambda L)^2 \sum_{j=0}^{r-1}\left(\e_j+\mathbb{E}\left[\left\|\nabla J\left(\theta_j\right)\right\|^2\right]\right)(1+\zeta)^{r-j}
\end{align*} 
By choosing $\zeta =\frac{1}{r}$, we have
\begin{align}\label{eq:upper_J}
\mathbb{E}\left[\left\|\nabla J_i\left(\theta_r\right)\right\|^2\right]  \leq e\mathbb{E}\left[\left\|\nabla J_i\left(\theta_0\right)\right\|^2\right]+2e(r+1)(\lambda L)^2 \sum_{j=0}^{r-1}\left(\e_j+\mathbb{E}\left[\left\|\nabla J\left(\theta_j\right)\right\|^2\right]\right)
\end{align} 
Summing up Eq.~\eqref{eq:local_drift_high} over $r=0,1,\cdots, R-1$ and then averaing Eq.~\eqref{eq:local_drift_high} over all $i \in N$, we have
\begin{align}
\sum_{r=0}^{R-1}\mathcal{M}_r &\le \sum_{r=0}^{R-1} \E\left[\frac{289}{144}\eta^2 (1-\beta)^2 \|u_r\|^2 +4 \eta^2 \beta^2\frac{1}{N}\sum_{i=1}^N\left\|\nabla J_i\left(\theta_r\right)\right\|^2 \right]\notag\\
& \le \sum_{r=0}^{R-1}\frac{289}{72}\eta^2(1-\beta)^2 \left(\e_{r-1}+\E[\|\nabla J(\theta_{r-1})\|^2] \right) \notag\\
& \stackrel{(b)}{+} 4 \eta^2 \beta^2 \sum_{r=0}^{R-1}\left(\frac{e}{N} \sum_{i=1}^N \mathbb{E}\left[\left\|\nabla J_i\left(\theta_0\right)\right\|^2\right]+2 e(r+1)(\lambda L)^2 \sum_{j=0}^{r-1}\left(\e_j+\mathbb{E}\left[\left\|\nabla J\left(\theta_j\right)\right\|^2\right]\right)\right)\\
& \le \frac{289}{72} \eta^2(1-\beta)^2 \sum_{r=0}^{R-1}\left(\e_{r-1}+\mathbb{E}\left[\left\|\nabla J\left(\theta_{r-1}\right)\right\|^2\right]\right) \notag\\
& +4 \eta^2 \beta^2\left(e R G_0+2 e(\lambda L R)^2 \sum_{r=0}^{R-2}\left(\e_r+\mathbb{E}\left[\left\|\nabla J\left(\theta_r\right)\right\|^2\right]\right)\right) \notag\\
& \stackrel{(c)}{\leq}  \frac{\beta^2}{288 e K^2 \widetilde{L_1}^2} \sum_{r=-1}^{R-2}\left(\e_r+\mathbb{E}\left[\left\|\nabla J\left(\theta_r\right)\right\|^2\right]\right)+4 \eta^2 \beta^2 e R G_0 .\notag
\end{align}
where (b) is due to the upper bound of $\mathbb{E}\left[\left\|\nabla J_i\left(\theta_r\right)\right\|^2\right]$ in Eq.\eqref{eq:upper_J} and (c) is due to the choice of $\eta$ such that $\eta^2 \left[\frac{289}{72}(1-\beta)^2+8 e(\lambda \beta L R)^2\right] \leq \frac{\beta^2}{288eK^2 \widetilde{L_1}^2}.$ 
\end{proof}

\vspace*{3em}
\subsection{Proof of Theorem~\ref{thm:fedsvrpg_m}}
\begin{theorem} (Complete version of Theorem~\ref{thm:fedsvrpg_m})
Under Assumptions~\ref{assume_policy}--\ref{assume_IS}, by setting $u_0=\frac{1}{N B} \sum_{i=1}^N \sum_{b=1}^B g_i\left(\tau_b^{(i)}|\theta_0\right)$ with $\left\{\tau^{(i)}_b\right\}_{b=1}^B \stackrel{iid}{\sim} p^{(i)}(\tau | \theta_0)$ and choosing $\beta =\min \left\{1,\left(\frac{N K \bar{L}^2 \Delta^2}{\sigma^4 R^2}\right)^{1 / 3}\right\}$,  $\lambda =\min \left\{\frac{1}{24 \bar{L}}, \sqrt{\frac{\beta N K}{162 \bar{L}^2}}\right\}$, $B=\left\lceil\frac{K}{R \beta^2}\right\rceil$, and  
$$
\eta K \bar{L}\lesssim \min \left\{\left(\frac{\bar{L} \Delta}{G_0 \lambda \bar{L} R}\right)^{1 / 2},\left(\frac{\beta}{N}\right)^{1 / 2},\left(\frac{\beta}{N K}\right)^{1 / 4}\right\}
$$ in Algorithm~\ref{algFedRL}, then the output of \textsc{FedSVRPG-M} after $R$ rounds satisfies:
\begin{align}
\frac{1}{R} \sum_{r=0}^{R-1} \mathbb{E}\left[\left\|\nabla J\left(\theta_r\right)\right\|^2\right] \lesssim\left(\frac{\bar{L} \Delta \sigma}{N K R}\right)^{2 / 3}+\frac{\bar{L}\Delta}{R},
\end{align}
where $\bar{L}:= \max\{L, \widetilde{L_1}, \widetilde{L_2}, \widetilde{L_3}\}$ and $L, \widetilde{L_1}, \widetilde{L_2}, \widetilde{L_3}$ are defined in Proposition~\ref{prop:lipschitz}, Lemma~\ref{lem:gradient_bias_SVRG} and Lemma~\ref{lem:drift_term_svrg}, respectively. 
\end{theorem}

\begin{proof} Based on Lemma~\ref{lem:gradient_bias_SVRG}, we have for any $r\ge 1$
\begin{align}
\Sigma_{r} &\le (1-\frac{8\beta}{9})\Sigma_{r-1} + \frac{\widetilde{L_1}^2}{\beta} \mathcal{D}_r + \frac{3\beta^2 \sigma^2}{NK} +18\lambda^2\frac{\widetilde{L_2}^2}{NK}\E\bigg\|\nabla J(\theta_{r-1}) \bigg\|^2 \notag\\
& \le (1-\frac{8\beta}{9})\Sigma_{r-1}  +18\lambda^2\frac{\widetilde{L_2}^2}{NK}\E\bigg\|\nabla J(\theta_{r-1}) \bigg\|^2 + \frac{3\beta^2 \sigma^2}{NK}\notag\\
& +\frac{\widetilde{L_1}^2}{\beta}\left(4e K^2\mathcal{M}_r + (16\eta^4 K^4L^2 + 8\eta^2 K)\right) \left(\beta^2\sigma^2 + 2 \widetilde{L_3}^2\E\Big\|\theta_{r-1} -\theta_{r}\Big\|^2 \right)
\end{align}
where the last inequality is due to Lemma~\ref{lem:drift_term_svrg}. When $r=0$, we have
\begin{align*}
\Sigma_{0} 
& \le (1-\beta)\Sigma_{-1} +\frac{3\beta^2 \sigma^2}{NK} +\frac{\widetilde{L_1}^2}{\beta}\left(4e K^2\mathcal{M}_0 + (16\eta^4 K^4L^2 + 8\eta^2 K)\right) \beta^2\sigma^2 
\end{align*}
Summing up the above equation over $r$ from $0$ to $R-1$, we have
\begin{align*}
\sum_{r=0}^{R-1} \e_r \leq & \left(1-\frac{8 \beta}{9}\right) \sum_{r=-1}^{R-2} \e_r+\frac{18(\lambda \widetilde{L_2})^2}{N K} \mathbb{E}\left[\sum_{r=0}^{R-2}\left\|\nabla J\left(\theta_r\right)\right\|^2\right]+\frac{3 \beta^2 \sigma^2}{N K} R \notag\\
& +\frac{\widetilde{L_1}^2}{\beta} \left(4 e K^2 \sum_{r=0}^{R-1} \mathcal{M}_r+8(\eta K)^2\left(2(\eta K L)^2+\frac{1}{K}\right)\left(R \beta^2 \sigma^2+2 L^2 \sum_{r=0}^{R-1} \mathbb{E}\left[\left\|\theta_r-\theta_{r-1}\right\|^2\right]\right)\right)
\end{align*}
By incorporating Lemma~\ref{lem:high_order_svrg} into the inequality above, we have
\begin{align}
\sum_{r=0}^{R-1} \e_r \leq & \left(1-\frac{8 \beta}{9}\right) \sum_{r=-1}^{R-2} \e_r+\frac{18(\lambda \widetilde{L_2})^2}{N K} \mathbb{E}\left[\sum_{r=0}^{R-2}\left\|\nabla J\left(\theta_r\right)\right\|^2\right]+\frac{3 \beta^2 \sigma^2}{N K} R \notag\\
& +\frac{\widetilde{L_1}^2}{\beta} 8(\eta K)^2\left(2(\eta K L)^2+\frac{1}{K}\right)\left(R \beta^2 \sigma^2+2 L^2 \sum_{r=0}^{R-1} \mathbb{E}\left[\left\|\theta_r-\theta_{r-1}\right\|^2\right]\right) \notag\\
&+\frac{\widetilde{L_1}^2}{\beta}4eK^2\left\{\frac{\beta^2}{288 e K^2 \widetilde{L_1}^2} \sum_{r=-1}^{R-2}\left(\e_r+\mathbb{E}\left[\left\|\nabla J\left(\theta_r\right)\right\|^2\right]\right)+4 \eta^2 \beta^2 e R G_0 \right\} \notag\\
&\le \left[1-\frac{8 \beta}{9} +\frac{\beta}{72}+\frac{32(\eta K\widetilde{L_1})^2}{\beta}(2(\eta K L)^2+\frac{1}{K})(\lambda L)^2\right] \sum_{r=-1}^{R-2} \e_r\notag\\
&+\left[\frac{18(\lambda \widetilde{L_2})^2}{N K} + \frac{32(\eta K\widetilde{L_1})^2}{\beta}(2(\eta K L)^2+\frac{1}{K})(\lambda L)^2 + \frac{\beta}{72}\right]\sum_{r=-1}^{R-2}\mathbb{E}\left[\left\|\nabla J\left(\theta_r\right)\right\|^2\right] \notag\\
&+\left[8\beta\widetilde{L_1}^2 (\eta K)^2(2(\eta K L)^2+\frac{1}{K})+\frac{3\beta^2}{NK}\right] R\sigma^2 + 16\beta(e\eta K \widetilde{L_1})^2 R G_0
\end{align}
Where the last inequality is derived by $\left\|\theta_r-\theta_{r-1}\right\|^2 \leq 2 \lambda^2\left(\left\|\nabla J\left(\theta_{r-1}\right)\right\|^2+\left\|u_r-\nabla J\left(\theta_{r-1}\right)\right\|^2\right)$. We require the following inequalities to hold, 
\begin{equation}
\begin{cases}
&\frac{32(\eta K\widetilde{L_1})^2}{\beta}(2(\eta K L)^2+\frac{1}{K})(\lambda L)^2 \le \frac{\beta}{18}\\
&8\widetilde{L_1}^2 (\eta K)^2(2(\eta K L)^2+\frac{1}{K}) \le \frac{\beta^2}{NK}\\
& \lambda \widetilde{L_2} \le \sqrt{\frac{\beta NK}{162}}.
    \end{cases}  
\end{equation}
Then, we have that 
\begin{align*}
\sum_{r=0}^{R-1} \e_r  &\le \left[1-\frac{8 \beta}{9} +\frac{\beta}{72}+\frac{\beta}{18}\right] \sum_{r=-1}^{R-2} \e_r
+\left[\frac{\beta}{9} + \frac{\beta}{18} + \frac{\beta}{72}\right]\sum_{r=-1}^{R-2}\mathbb{E}\left[\left\|\nabla J\left(\theta_r\right)\right\|^2\right] \notag\\
&+\left[\frac{\beta^2}{NK}+\frac{3\beta^2}{NK}\right] R\sigma^2 + 16\beta(e\eta K \widetilde{L_1})^2 R G_0 \\
& \le (1-\frac{7 \beta}{9})\sum_{r=-1}^{R-2} \e_r + \frac{2\beta}{9}\sum_{r=-1}^{R-2}\mathbb{E}\left[\left\|\nabla J\left(\theta_r\right)\right\|^2\right] +\frac{4R\beta^2\sigma^2}{NK} + 16\beta(e \eta K \widetilde{L_1})^2 R G_0 
\end{align*}
After some rearrangement, we have
\begin{align*}
\sum_{r=0}^{R-1} \e_r  &\le \frac{9}{7\beta} \e_{-1} +\frac{2}{7}\sum_{r=-1}^{R-2}\mathbb{E}\left[\left\|\nabla J\left(\theta_r\right)\right\|^2\right] +\frac{36 R\beta \sigma^2}{7NK} +\frac{144}{7}(e\eta K \widetilde{L_1})^2 R G_0 
\end{align*}
Based on Lemma~\ref{lem:function_decay}, we have
\begin{align*}
\frac{1}{\lambda}\E[J(\theta_R) -J(\theta_0)] \ge \frac{2}{7}\sum_{r=0}^{R-1}\mathbb{E}\left[\left\|\nabla J\left(\theta_r\right)\right\|^2\right] -\frac{1}{35\beta}\e_{-1}-\frac{39 R\beta \sigma^2}{14NK} -\frac{78}{7}(e\eta K \widetilde{L_1})^2 R G_0 
\end{align*}
Notice that $u_0=\frac{1}{N B} \sum_i \sum_{b=1}^B g_i\left(\tau_b^{(i)}|\theta_0\right)$ implies $\e_{-1}=\E\| u_0 -\nabla J(\theta_0)\|^2 \leq \frac{\sigma^2}{N B} \leq \frac{\beta^2 \sigma^2 R}{N K}$.  Define $\bar{L}:= \max\{L, \widetilde{L_1}, \widetilde{L_2}, \widetilde{L_3}\}$ and after some rearrangement, we have
\begin{align}
\frac{1}{R} \sum_{r=0}^{R-1} \mathbb{E}\left[\left\|\nabla J\left(\theta_r\right)\right\|^2\right] & \lesssim \frac{\bar{L} \Delta}{\lambda \bar{L} R}+\frac{\e_{-1}}{\beta R}+(\eta K \widetilde{L_1})^2 G_0+\frac{\beta \sigma^2}{N K} \notag\\
& \stackrel{(a)}{\lesssim} \frac{\bar{L} \Delta}{\lambda \bar{L} R}+\frac{\beta \sigma^2}{N K} \notag\\
& \stackrel{(b)}{\lesssim} \frac{\bar{L} \Delta}{R}+\frac{\bar{L} \Delta}{\sqrt{\beta N K}}+\frac{\beta \sigma^2}{N K} \notag\\
&\stackrel{(c)}{\lesssim} \frac{\bar{L} \Delta}{R}+\left(\frac{\bar{L} \Delta \sigma}{N K R}\right)^{2 / 3}\notag
\end{align}
where $(a)$ is due to the fact $\eta K \bar{L} \lesssim \left(\frac{\bar{L}\Delta}{G_0 \lambda L R}\right)^{\frac{1}{2}};$ For (b), it holds because $\lambda \bar{L} \le \min\{\frac{1}{24}, \sqrt{\frac{\beta NK}{162}}\};$ For (c), it holds because $\beta=\min \left\{1,\left(\frac{N K \bar{L}^2 \Delta^2}{\sigma^4 R^2}\right)^{1 / 3}\right\}.$
\end{proof}

\newpage
\section{Federated Hessian Aided Policy Gradient with Momentum }
According to the updating rule of \textsc{FEDHAPG-M}, we can rewrite $\Lambda_{r,k}^{(i)}$ as
\begin{align}
\Lambda_{r,k}^{(i)}= & \left(\nabla \log p^{(i)}\left(\tau_{r,k}^{(i)} \mid \theta_{r,k}^{(i)}(\alpha)\right)^T v_{r,k}^{(i)}\right) \nabla \Phi_i\left(\tau_{r,k}^{(i)} \mid \theta_{r,k}^{(i)}(\alpha)\right)  +\nabla^2 \Phi_i\left(\tau_{r,k}^{(i)} \mid \theta_{r,k}^{(i)}(\alpha)\right)v_{r,k}^{(i)}
\end{align}
where $
\Phi_i(\tau \mid \theta)=\sum_{h=0}^{H-1} \sum_{i=h}^{H-1} \lambda^i \mathcal{R}^{(i)}\left(s_i, a_i\right) \log \pi_\theta\left(a_h, s_h\right)
$ and $v_{r,k}^{(i)} = \theta_{r,k}^{(i)} -\theta_{r-1}$. Note that $$
\mathbb{E}_{\alpha \sim U[0,1], \tau \sim p^{(i)}\left(\tau \mid \theta_{r,k}^{(i)}(\alpha)\right)}\left[\Lambda_{r,k}^{(i)}\right]=\nabla J\left(\theta_{r,k}^{(i)}\right)-\nabla J\left(\theta_{r-1}\right).
$$
Moreover, we have $
\Lambda_{r,k}^{(i)}:=\hat{\nabla}_i^2\left(\theta_{r,k}^{(i)}(\alpha), \tau_{r,k}^{(i)}\right) v_{r,k}^{(i)} 
$ 
where
$$
\begin{aligned}
\hat{\nabla}_i^2\left(\theta_{r,k}^{(i)}(\alpha), \tau_{r,k}^{(i)}\right)= & \nabla \Phi_i\left(\tau_{r,k}^{(i)} \mid \theta_{r,k}^{(i)}(\alpha)\right) \nabla \log p^{(i)}\left(\tau_{r,k}^{(i)} \mid \theta_{r,k}^{(i)}(\alpha)\right)^T \\
& +\nabla^2 \Phi_i\left(\tau_{r,k}^{(i)} \mid \theta_{r,k}^{(i)}(\alpha)\right) .
\end{aligned}
$$ and
$\mathbb{E}_{\tau \sim p^{(i)}\left(\tau \mid \theta_{r,k}^{(i)}(\alpha)\right)}\left[\hat{\nabla}^2\left(\theta_{r,k}^{(i)}(\alpha), \tau\right)\right]=\nabla^2 J_i\left(\theta_{r,k}^{(i)}(\alpha)\right).$

\vspace*{3em}
\begin{prop}\label{prop: second_gradient} (Lemma 4.1 in~\cite{shen2019hessian}) Under Assumption~\ref{assume_policy}, we have for all $\theta$ and $i\in [N]$
$$
\left\|\hat{\nabla}_i^2(\theta, \tau)\right\|^2 \leq \frac{H^2 G^4 R_{\max}^2+M^2 R^2}{(1-\lambda)^4}=\widetilde{L_4}^2 .
$$
where $\tau$ is a trajectory sampled according to $p^{(i)}(\tau | \theta)$.
\end{prop}
\vspace*{3em}
\begin{lemma}\label{lem:dreasing_HA}
If the step-size satisfies $\lambda \le \sqrt{\frac{\beta NK}{72\widetilde{L_4}^2}},$ we have
\begin{align}
\e_r \le (1-\frac{8\beta}{\beta}) \e_{r-1}+\frac{2L^2 +4\widetilde{L_4}^2}{\beta}\mathcal{D}_r + \frac{2\beta^2 \sigma^2}{NK} +\frac{8\lambda^2\widetilde{L_4}^2}{NK}\E\left\|\nabla J(\theta_{r-1}) \right\|^2
\end{align}
holds for $r \ge 1.$ When $r=0,$ we have 
\begin{align}
\e_r \le (1-\frac{8\beta}{\beta}) \e_{r-1}+\frac{2L^2 +4\widetilde{L_4}^2}{\beta}\mathcal{D}_r + \frac{2\beta^2 \sigma^2}{NK}. 
\end{align}
\end{lemma}
\begin{proof}
\begin{align} \label{eq:HAPG_gradient_bias}
\e_r &= \E\left[\left\lVert  u_{r+1} -\nabla J(\theta_r)  \right\rVert^2\right] \notag\\
& = \E \left[ \Big\lVert \frac{1}{NK}\sum_{i,k}\beta w^{(i)}\left(\tau_{r,k}^{(i)} \mid  \theta_{r,k}^{(i)},\theta_{r,k}^{(i)}(\alpha)\right)g_i\left(\tau_{r,k}^{(i)}\mid \theta_{r,k}^{(i)}\right)+\left(1-\beta\right)\left[u_r+\Lambda_{r,k}^{(i)}\right]-\nabla J(\theta_{r}) \Big\rVert^2\right] \notag\\
& = \E\Bigg[\bigg\| (1-\beta)(u_r -\nabla J(\theta_{r-1}))\notag\\
&+\frac{1}{NK}\sum_{i,k}\left\{\beta w^{(i)}\left(\tau_{r,k}^{(i)} \mid  \theta_{r,k}^{(i)},\theta_{r,k}^{(i)}(\alpha)\right)g_i\left(\tau_{r,k}^{(i)}\mid \theta_{r,k}^{(i)}\right) +(1-\beta)\left(\Lambda_{r,k}^{(i)}+\nabla J (\theta_{r-1}\right) -\nabla J(\theta_{r,k}^{(i)})\right\}\notag\\
&+ \frac{1}{NK}\sum_{i,k}\left[\nabla J(\theta_{r,k}^{(i)})-\nabla J(\theta_r)\right]\bigg\|^2 \Bigg]\notag\\
& = (1-\beta)^2 \e_{r-1} + \underbrace{\E\left\|\frac{1}{NK}\sum_{i,k}\left[\nabla J(\theta_{r,k}^{(i)})-\nabla J(\theta_r)\right]\right\|^2}_{H_1}\notag\\
&+\underbrace{2\E\left[\Big\langle(1-\beta)(u_r -\nabla J(\theta_{r-1})), \frac{1}{NK}\sum_{i,k}\left[\nabla J_i(\theta_{r,k}^{(i)}) -\nabla J (\theta_r)\right]\Big\rangle \right]}_{H_2}\notag\\
& + \underbrace{\E\left\| \frac{1}{NK}\sum_{i,k}\left\{\beta w^{(i)}\left(\tau_{r,k}^{(i)} \mid  \theta_{r,k}^{(i)},\theta_{r,k}^{(i)}(\alpha)\right)g_i\left(\tau_{r,k}^{(i)}\mid \theta_{r,k}^{(i)}\right) +(1-\beta)\left(\Lambda_{r,k}^{(i)}+\nabla J (\theta_{r-1})\right) -\nabla J(\theta_{r,k}^{(i)})\right\}\right\|^2}_{H_3} 
\end{align}

To precede, we bound $H_1$ as
\begin{align}\label{eq:H_1}
H_1 &=\E\left\|\frac{1}{NK}\sum_{i,k}\left[\nabla J(\theta_{r,k}^{(i)})-\nabla J(\theta_r)\right]\right\|^2\notag\\
& \le \frac{L^2}{NK}\sum_{i,k}\E\Big\lVert \theta_{r,k}^{(i)} -\theta_r\Big\rVert^2 =L^2 \mathcal{D}_r
\end{align}

Using Young's inequality to bound $H_2$, we have 
\begin{align}\label{eq:H_2}
H_2 &\le \beta(1-\beta)^2\E\Big\lVert u_r -\nabla J(\theta_{r-1})\Big\rVert^2 + \frac{1}{\beta}\E\Big\lVert \frac{1}{NK}\sum_{i,k}\nabla J_i(\theta_{r,k}^{(i)}) -\nabla J (\theta_r)\Big\rVert^2\notag\\
&\le \beta(1-\beta)^2 \Sigma_{r-1} + \frac{L^2}{\beta}\underbrace{\frac{1}{NK} \sum_{i,k} \E\left\|\theta_{r,k}^{(i)}-\theta_r \right\|^2}_{\mathcal{D}_r}
\end{align}

For $H_3$, we bound it as 
\begin{align}\label{eq:H_3}
H_3 &= \E\left\| \frac{1}{NK}\sum_{i,k}\left\{\beta w^{(i)}\left(\tau_{r,k}^{(i)} \mid  \theta_{r,k}^{(i)},\theta_{r,k}^{(i)}(\alpha)\right)g_i\left(\tau_{r,k}^{(i)}\mid \theta_{r,k}^{(i)}\right) +(1-\beta)\left(\Lambda_{r,k}^{(i)}+\nabla J (\theta_{r-1})\right) -\nabla J(\theta_{r,k}^{(i)})\right\}\right\|^2 \notag\\
& \le 2\beta^2 \frac{\sigma^2}{NK} +2(1-\beta)^2\frac{1}{N^2K^2}\sum_{i,k}\E\left\|\Lambda_{r,k}^{(i)}+\nabla J (\theta_{r-1}) -\nabla J(\theta_{r,k}^{(i)})\right\|^2\notag\\
& \stackrel{(a)}{\le} \frac{2\beta^2\sigma^2}{NK} +2(1-\beta)^2 \frac{1}{N^2K^2}\sum_{i,k}\E\left\|\Lambda_{r,k}^{(i)}\right\|^2 \notag\\
& \stackrel{(b)}{=}\frac{2\beta^2\sigma^2}{NK} +2(1-\beta)^2 \frac{1}{N^2K^2}\sum_{i,k}\E\left\|\hat{\nabla}^2\left(\theta_{r,k}^{(i)}, \tau_{r,k}^{(i)}\right) v_{r,k}^{(i)}\right\|^2 \stackrel{(b)}{\le} \frac{2\beta^2\sigma^2}{NK} +2(1-\beta)^2\frac{1}{N^2K^2}\sum_{i,k}\widetilde{L_4}^2 \E\left\| v_{r,k}^{(i)}\right\|^2 \notag\\
& \le \frac{2\beta^2\sigma^2}{NK} +4(1-\beta)^2\widetilde{L_4}^2\underbrace{\frac{1}{NK}\sum_{i,k}\E\left\| \theta_{r,k}^{(i)}-\theta_r\right\|^2}_{\mathcal{D}_r} +4(1-\beta)^2\frac{\widetilde{L_4}^2}{NK} \E\left\|\theta_{r-1}-\theta_r \right\|^2
\end{align}
where we use the fact that $\E[\| X -\E[X]\|^2]\le \E[\|X\|^2]$ for (a); for (b), it holds due to Proposition~\ref{prop: second_gradient}.

Plugging the upper bound of $H_1$ (Eq.~\eqref{eq:H_1}), $H_2$(Eq.~\eqref{eq:H_2}) and $H_3$ (Eq.~\eqref{eq:H_3})into Eq.\eqref{eq:HAPG_gradient_bias}, we have
\begin{align}
\e_r &\le (1-\beta)\e_{r-1} +\frac{2L^2 +4\widetilde{L_4}^2}{\beta}\mathcal{D}_r + \frac{2\beta^2 \sigma^2}{NK} + 4\frac{\widetilde{L_4}^2}{NK}\E\left\|\theta_{r-1}-\theta_r \right\|^2 \notag\\
&=(1-\beta)\e_{r-1} +\frac{2L^2 +4\widetilde{L_4}^2}{\beta}\mathcal{D}_r + \frac{2\beta^2 \sigma^2}{NK} + 4\frac{\lambda^2\widetilde{L_4}^2}{NK}\E\left\|u_r \right\|^2 \notag\\
& \le (1-\beta)\e_{r-1} +\frac{2L^2 +4\widetilde{L_4}^2}{\beta}\mathcal{D}_r + \frac{2\beta^2 \sigma^2}{NK} + 8\frac{\lambda^2\widetilde{L_4}^2}{NK}\E\left\|u_r -\nabla J(\theta_{r-1}) \right\|^2 +8\frac{\lambda^2\widetilde{L_4}^2}{NK}\E\left\|\nabla J(\theta_{r-1}) \right\|^2 \notag\\
& \stackrel{(a)}{\le} (1-\frac{8\beta}{\beta}) \e_{r-1}+\frac{2L^2 +4\widetilde{L_4}^2}{\beta}\mathcal{D}_r + \frac{2\beta^2 \sigma^2}{NK} +\frac{8\lambda^2\widetilde{L_4}^2}{NK}\E\left\|\nabla J(\theta_{r-1}) \right\|^2 
\end{align}
where (a) is due to the choice of $\lambda$ such that $\frac{8\lambda^2\widetilde{L_4}^2}{NK} \le \frac{\beta}{9}$, which holds when $\lambda \le \sqrt{\frac{\beta NK}{72\widetilde{L_4}^2}}.$
\end{proof}

\vspace*{3em}
\begin{lemma}\label{lem:bounding_drift_HA}
(Bounding drift-term) If the local step-size satisfies $\eta \le \min\{\frac{L}{32e^2\widetilde{L_4}^2 K},\frac{1}{KL}\}$, the drift-term can be upper bounded as: 
\begin{align*}
\mathcal{D}_r \le 4e K^2 \mathcal{M}_r + (16\eta^4 K^4L^2 + 8\eta^2 K) \left(\beta^2\sigma^2 + 2 \widetilde{L_4}^2\E\Big\|\theta_{r-1} -\theta_{r}\Big\|^2 \right)
\end{align*}
\end{lemma}
\begin{proof}
Define $c_{r,k}^{(i)}:= -\eta \left(\nabla J_i(\theta_{r,k}^{(i)}) +(1-\beta)(u_r -\nabla J_{i}(\theta_{r-1}))\right)$. For any $1\le j \le k-1 \le K-2,$ we have:
\begin{align}\label{eq:variance_decomposition}
\E\left\|c_{r,j}^{(i)} -c_{r,j-1}^{(i)} \right\|^2 &\le \eta^2 L^2 \E\left\|\theta_{r,j}^{(i)} -\theta_{r,j-1}^{(i)} \right\|^2\notag\\
&= \eta^2 L^2\left(\E\left\|c_{r,j-1}^{(i)}\right\|^2 + \mathbb{E}\left[\operatorname{Var}\left[\theta^{(i)}_{r, j}-\theta^{(i)}_{r, j-1} \mid \mathcal{F}^{(i)}_{r, j-1}\right]\right]\right).
\end{align}
where we use the bias-variance decomposition in the last inequality. To precede, we bound the variance term as:
\begin{align}\label{eq:HAPG_upper_variance}
&\mathbb{E}\left[\operatorname{Var}\left[\theta^{(i)}_{r, j}-\theta^{(i)}_{r, j-1} \mid \mathcal{F}^{(i)}_{r, j-1}\right]\right] \notag\\
&=\eta^2\E\bigg\|\beta\left[w^{(i)}\left(\tau_{r,j-1}^{(i)} \mid  \theta_{r,j-1}^{(i)},\theta_{r,j-1}^{(i)}(\alpha)\right)g_i\left(\tau_{r,j-1}^{(i)}\mid \theta_{r,j-1}^{(i)}\right)-\nabla J_i(\theta_{r,j-1}^{(i)})\right]\notag\\
&+\left(1-\beta\right)\left[ \Lambda_{r,j-1}^{(i)}-\nabla J_i(\theta_{r,j-1}^{(i)})+\nabla J_i (\theta_{r-1})\right]\bigg\|^2\notag\\
& \le 2\eta^2 \beta^2\sigma^2  + 2 \eta^2 (1-\beta)^2 \E\bigg\|\Lambda_{r,j-1}^{(i)}-\nabla J_i(\theta_{r,j-1}^{(i)})+\nabla J_i (\theta_{r-1})\bigg\|^2\notag\\
& \stackrel{(a)}{\le} 2\eta^2 \beta^2\sigma^2  + 2 \eta^2 (1-\beta)^2 \E\bigg\|\Lambda_{r,j-1}^{(i)}\bigg\|^2 \notag\\
& \le 2\eta^2 \beta^2\sigma^2  + 2 \eta^2 (1-\beta)^2 \E\left\|\hat{\nabla}_i^2\left(\theta_{r,j-1}^{(i)}, \tau_{r,j-1}^{(i)}\right) v_{r,j-1}^{(i)}\right\|^2 \notag\\
& \le 2\eta^2 \beta^2\sigma^2  + 4\eta^2(1-\beta)^2\widetilde{L_4}^2\E\left\| \theta_{r,j-1}^{(i)}-\theta_r\right\|^2 +4\eta^2(1-\beta)^2\widetilde{L_4}^2\E\left\|\theta_{r-1}-\theta_r \right\|^2
\end{align}
where we use the fact that $\E[\| X -\E[X]\|^2]\le \E[\|X\|^2]$ for (a). Plugging the upper bound of variance into Eq.\eqref{eq:variance_decomposition}, we have
\begin{align}
\E\left\|c_{r,j}^{(i)} -c_{r,j-1}^{(i)} \right\|^2
& \le \eta^2 L^2\bigg(\E\left\|c_{r,j-1}^{(i)}\right\|^2 +2\eta^2\beta^2\sigma^2 + 4\eta^2 \widetilde{L_4}^2\E\Big\|\theta_{r,j-1}^{(i)} -\theta_{r}\Big\|^2+ 4\eta^2 \widetilde{L_4}^2\E\Big\|\theta_{r-1} -\theta_{r}\Big\|^2\bigg). \notag
\end{align}

Then for any $1\le j \le k-1 \le K-2$, we have 
\begin{align}
&\E\left\|c_{r,j}^{(i)} \right\|^2 \le (1+\frac{1}{q})\E\left\|c_{r,j-1}^{(i)} \right\|^2 + (1+q)\E\left\|c_{r,j}^{(i)} -c_{r,j-1}^{(i)} \right\|^2 \notag\\
& \le (1+\frac{2}{q})\E\left\|c_{r,j-1}^{(i)} \right\|^2 + (1+q)\eta^2 L^2 \left( 2\eta^2\beta^2\sigma^2 +4\eta^2 \widetilde{L_4}^2\E\Big\|\theta_{r,j-1}^{(i)} -\theta_{r}\Big\|^2+4\eta^2 \widetilde{L_4}^2\E\Big\|\theta_{r-1} -\theta_{r}\Big\|^2 \right) 
\end{align}
where we use the fact that $\eta L \le \frac{1}{K}\le\frac{1}{q+1}$ and let $q=k-1$. By unrolling this recurrence, for any $1\le j \le k-1 \le K-2$, we have
\begin{align}
\E\left\|c_{r,j}^{(i)} \right\|^2 &\le (1+\frac{2}{k-1})^{j}\E\left\|c_{r,0}^{(i)} \right\|^2  + k\eta^2 L^2\sum_{i=0}^{j-1}(2\eta^2\beta^2\sigma^2 +4\eta^2 \widetilde{L_4}^2\E\Big\|\theta_{r-1} -\theta_{r}\Big\|^2)\Pi_{j^{\prime}=i+1}^{j-1}(1+\frac{2}{k-1}) \notag\\
& + k\eta^2 L^2\sum_{s=0}^{j-1}(4\eta^2 \widetilde{L_4}^2\E\Big\|\theta_{r,s}^{(i)} -\theta_{r}\Big\|^2)\Pi_{j^{\prime}=s+1}^{j-1}(1+\frac{2}{k-1}) \notag\\
& \le (1+\frac{2}{k-1})^{k-1}\E\left\|c_{r,0}^{(i)} \right\|^2  + k\eta^2 L^2\sum_{i=0}^{k-1}(2\eta^2\beta^2\sigma^2 +4\eta^2 \widetilde{L_4}^2\E\Big\|\theta_{r-1} -\theta_{r}\Big\|^2)(1+\frac{2}{k-1})^{k-1} \notag\\
& + k\eta^2 L^2\sum_{j^{\prime}=0}^{j-1}(4\eta^2 \widetilde{L_4}^2\E\Big\|\theta_{r,j^{\prime}}^{(i)} -\theta_{r}\Big\|^2)(1+\frac{2}{k-1})^{k-1}
\end{align}
Based on the inequality $(1+\frac{2}{K-1}^{k-1})\le e^2 \le 8$, we have 
\begin{align}\label{eq:HAPG_upper_c}
\E\left\|c_{r,j}^{(i)} \right\|^2 &\le e^2\E\left\|c_{r,0}^{(i)} \right\|^2 + 8k^2\eta^4L^2 \left(2\beta^2\sigma^2 + 4 \widetilde{L_4}^2\E\Big\|\theta_{r-1} -\theta_{r}\Big\|^2 \right) + 4e^2 k\eta^4 L^2 \widetilde{L_4}^2\sum_{j^{\prime}=0}^{j-1}\E\Big\|\theta_{r,j^{\prime}}^{(i)} -\theta_{r}\Big\|^2
\end{align}
By Lemma A.3, we have
\begin{align}\label{eq:HAPG_drift_deviation}
\E\left\|\theta_{r,k}^{(i)} -\theta_r \right\|^2 &\le 2 \E\left\| \sum_{j=0}^{k-1} c_{r,j}^{(i)}\right\|^2 + 
2 \sum_{j=0}^{k-1} \mathbb{E}\left[\operatorname{Var}\left[\theta_{r, j+1}^{(i)}-\theta_{r, j}^{(i)} \mid \mathcal{F}_{r, j}^{(i)}\right]\right] \notag\\
&\stackrel{(a)}{\leq} 2 k \sum_{j=0}^{k-1} \mathbb{E}\left\|  c_{r,j}^{(i)}\right\|^2+2 \sum_{j=0}^{k-1}\left(2 \beta^2 \eta^2 \sigma^2+4\eta^2 \widetilde{L_4}^2\E\Big\|\theta_{r,j}^{(i)} -\theta_{r}\Big\|^2+4\eta^2 \widetilde{L_4}^2\E\Big\|\theta_{r-1} -\theta_{r}\Big\|^2\right)
\end{align}
where $(a)$ is due to Eq.\eqref{eq:HAPG_upper_variance}. Plugging Eq.\eqref{eq:HAPG_upper_c} into Eq.\eqref{eq:HAPG_drift_deviation}, we have
\begin{align}
&\E\left\|\theta_{r,k}^{(i)} -\theta_r \right\|^2 \le\notag\\
& 2 k \sum_{j=0}^{k-1}\bigg\{e^2 \E\left\|c_{r,0}^{(i)} \right\|^2 + 8k^2\eta^4L^2 \left(2\beta^2\sigma^2 + 4 \widetilde{L_4}^2\E\Big\|\theta_{r-1} -\theta_{r}\Big\|^2 \right) + 4e^2 k\eta^4 L^2 \widetilde{L_4}^2\sum_{j^{\prime}=0}^{j-1}\E\Big\|\theta_{r,j^{\prime}}^{(i)} -\theta_{r}\Big\|^2\bigg\}\notag\\
&+2 \sum_{j=0}^{k-1}\left(2 \beta^2 \eta^2 \sigma^2+4\eta^2 \widetilde{L_4}^2\E\Big\|\theta_{r,j}^{(i)} -\theta_{r}\Big\|^2+4\eta^2 \widetilde{L_4}^2\E\Big\|\theta_{r-1} -\theta_{r}\Big\|^2\right)
\end{align}
Summing up the above equation over $k=0,\cdots, K-1,$ we have
\begin{align}
\sum_{k=0}^{K-1}\E\left\|\theta_{r,k}^{(i)} -\theta_r \right\|^2 &\le \sum_{k=0}^{K-1} \left\{2k^2 e^2 \E\left\|c_{r,0}^{(i)} \right\|^2 + 16k^4\eta^4L^2 \left(2\beta^2\sigma^2 + 4 \widetilde{L_4}^2\E\Big\|\theta_{r-1} -\theta_{r}\Big\|^2 \right)\right\} \notag\\
& +\sum_{k=0}^{K-1}8e^2 k^2 \eta^4 L^2 \widetilde{L_4}^2\sum_{j=0}^{k-1}\sum_{j^{\prime}=0}^{j-1}\E\Big\|\theta_{r,j^{\prime}}^{(i)} -\theta_{r}\Big\|^2 \notag\\
& +\sum_{k=0}^{K-1} \left(4k \beta^2 \eta^2 \sigma^2 + 8k\eta^2 \widetilde{L_4}^2\E\Big\|\theta_{r-1} -\theta_{r}\Big\|^2+ 8\eta^2 \widetilde{L_4}^2\sum_{j=0}^{k-1}\E\Big\|\theta_{r,j}^{(i)} -\theta_{r}\Big\|^2 \right)\notag\\
&\le 2e K^3 \E\left\|c_{r,0}^{(i)} \right\|^2 + (8\eta^4 K^5L^2 + 4\eta^2 K^2) \left(\beta^2\sigma^2 + 2 \widetilde{L_4}^2\E\Big\|\theta_{r-1} -\theta_{r}\Big\|^2 \right) \notag\\
& +K^2\sum_{k=0}^{K-1}8e^2  \eta^4 L^2 \widetilde{L_4}^2\sum_{j=0}^{K-1}\sum_{j^{\prime}=0}^{K-1}\E\Big\|\theta_{r,j^{\prime}}^{(i)} -\theta_{r}\Big\|^2 +\sum_{k=0}^{K-1}8\eta^2 \widetilde{L_4}^2\sum_{j=0}^{K-1}\E\Big\|\theta_{r,j}^{(i)} -\theta_{r}\Big\|^2 \notag\\
& =2e K^3 \E\left\|c_{r,0}^{(i)} \right\|^2 + (8\eta^4 K^5L^2 + 4\eta^2 K^2) \left(\beta^2\sigma^2 + 2 \widetilde{L_4}^2\E\Big\|\theta_{r-1} -\theta_{r}\Big\|^2 \right) \notag\\
& +(8e^2\eta^4 K^4L^2 \widetilde{L_4}^2+8\eta^2 \widetilde{L_4}^2K)\sum_{j=0}^{K-1}\E\Big\|\theta_{r,j}^{(i)} -\theta_{r}\Big\|^2
\end{align}
With the choice of step-size $\eta$ satisfying $8e^2\eta^4 K^4L^2 \widetilde{L_4}^2+8\eta^2 \widetilde{L_4}^2K \le \frac{1}{2},$ after some rearrangement, we have
\begin{align}
\frac{1}{2K}\sum_{k=0}^{K-1}\E\left\|\theta_{r,k}^{(i)} -\theta_r \right\|^2 \le 2e K^2 \E\left\|c_{r,0}^{(i)} \right\|^2 + (8\eta^4 K^4L^2 + 4\eta^2 K) \left(\beta^2\sigma^2 + 2 \widetilde{L_4}^2\E\Big\|\theta_{r-1} -\theta_{r}\Big\|^2 \right)
\end{align}
In summary, we can bound the drift-term as 
\begin{align}
\mathcal{D}_r \le 4e K^2 \mathcal{M}_r + (16\eta^4 K^4L^2 + 8\eta^2 K) \left(\beta^2\sigma^2 + 2 \widetilde{L_4}^2\E\Big\|\theta_{r-1} -\theta_{r}\Big\|^2 \right)
\end{align}
\end{proof}

\vspace*{3em}
\begin{lemma}\label{lem:high_order_HAPG}
If $\lambda L \le \frac{1}{24}$ and $\eta^2 \left[\frac{289}{72}(1-\beta)^2+8 e(\lambda \beta L R)^2\right] \leq \frac{\beta^2}{288eK^2 \left(2L^2+\widetilde{4L_4}^2\right)},$ we have
\begin{align}
\sum_{r=0}^{R-1}\mathcal{M}_r =\frac{1}{N} \sum_{r=0}^{R-1}\sum_{i=1}^N\E\left\|c_{r,0}^{(i)} \right\|^2\le \frac{\beta^2}{288 e K^2 \left(2L^2+\widetilde{4L_4}^2\right)} \sum_{r=-1}^{R-2}\left(\e_r+\mathbb{E}\left[\left\|\nabla J\left(\theta_r\right)\right\|^2\right]\right)+4 \eta^2 \beta^2 e R G_0 .
\end{align}
where $G_0: =\frac{1}{N} \sum_{i=1}^N \mathbb{E}\left[\left\|\nabla J_i\left(\theta_0\right)\right\|^2\right].$
\end{lemma}

\begin{proof}
The proof is the same as that of Lemma~\ref{lem:high_order_svrg}.
\end{proof}

\vspace*{3em}
\subsection{Proof of Theorem~\ref{thm:fedhapg}}
\begin{theorem} (Complete version of Theorem~\ref{thm:fedhapg})
Under Assumption~\ref{assume_policy}--\ref{assume_IS},  by setting $u_0=\frac{1}{N B} \sum_{i=1}^N \sum_{b=1}^B g_i\left(\tau_b^{(i)}|\theta_0\right)$ with $\left\{\tau^{(i)}_b\right\}_{b=1}^B \stackrel{iid}{\sim} p^{(i)}(\tau | \theta_0)$ and choosing $\beta =\min \left\{1,\left(\frac{N K \hat{L}^2 \Delta^2}{\sigma^4 R^2}\right)^{1 / 3}\right\}$,  $\lambda =\min \left\{\frac{1}{24 \hat{L}}, \sqrt{\frac{\beta N K}{72 \hat{L}^2}}\right\}$, $B=\left\lceil\frac{K}{R \beta^2}\right\rceil$, and  
$$
\eta K \hat{L}\lesssim \min \left\{\left(\frac{\hat{L} \Delta}{G_0 \lambda \hat{L} R}\right)^{1 /2}, \left(\frac{\beta}{N}\right)^{1 /2}, \left(\frac{\beta}{N K}\right)^{1/4}\right\}
$$ in Algorithm~\ref{algFedHAPG}, then the output of \textsc{FedHAPG-M} after $R$ rounds satisfies
\begin{align}
\frac{1}{R} \sum_{r=0}^{R-1} \mathbb{E}\left[\left\|\nabla J\left(\theta_r\right)\right\|^2\right] \lesssim\left(\frac{\hat{L} \Delta \sigma}{N K R}\right)^{2 / 3}+\frac{\hat{L}\Delta}{R}
\end{align}
where $\hat{L}:=\sqrt{2L^2+\widetilde{4L_4}^2}$ and $L, \widetilde{L_4}$ are defined in Proposition~\ref{prop:lipschitz} and Proposition~\ref{prop: second_gradient}, respectively.
\end{theorem}
\begin{proof} Based on Lemma~\ref{lem:dreasing_HA}, we have for any $r\ge 1$
\begin{align}
\e_r &\le (1-\frac{8\beta}{\beta}) \e_{r-1}+\frac{2L^2 +4\widetilde{L_4}^2}{\beta}\mathcal{D}_r + \frac{2\beta^2 \sigma^2}{NK} +\frac{8\lambda^2\widetilde{L_4}^2}{NK}\E\left\|\nabla J(\theta_{r-1}) \right\|^2\notag\\
& \le (1-\frac{8\beta}{\beta}) \e_{r-1} + \frac{2\beta^2 \sigma^2}{NK} +\frac{8\lambda^2\widetilde{L_4}^2}{NK}\E\left\|\nabla J(\theta_{r-1}) \right\|^2\\
&+\frac{2L^2 +4\widetilde{L_4}^2}{\beta}\left[4e K^2 \mathcal{M}_r + (16\eta^4 K^4L^2 + 8\eta^2 K) \left(\beta^2\sigma^2 + 2 \widetilde{L_4}^2\E\Big\|\theta_{r-1} -\theta_{r}\Big\|^2 \right)\right]
\end{align}
where the last inequality is due to Lemma~\ref{lem:bounding_drift_HA}. When $r=0$, we have
\begin{align*}
\Sigma_{0} 
& \le (1-\beta)\Sigma_{-1} +\frac{2\beta^2 \sigma^2}{NK} +\frac{2L^2 +4\widetilde{L_4}^2}{\beta}\left[4e K^2\mathcal{M}_0 + (16\eta^4 K^4L^2 + 8\eta^2 K)\right] \beta^2\sigma^2 
\end{align*}
Summing up the above equation over $r$ from $0$ to $R-1$, we have
\begin{align*}
\sum_{r=0}^{R-1} \e_r \leq & \left(1-\frac{8 \beta}{9}\right) \sum_{r=-1}^{R-2} \e_r+\frac{8(\lambda \widetilde{L_4})^2}{N K} \mathbb{E}\left[\sum_{r=0}^{R-2}\left\|\nabla J\left(\theta_r\right)\right\|^2\right]+\frac{2 \beta^2 \sigma^2}{N K} R \notag\\
& +\frac{2L^2 +4\widetilde{L_4}^2}{\beta} \left[4 e K^2 \sum_{r=0}^{R-1} \mathcal{M}_r+8(\eta K)^2(2(\eta K L)^2+\frac{1}{K})\right]\left(R \beta^2 \sigma^2+2 L^2 \sum_{r=0}^{R-1} \mathbb{E}\Big\lVert\theta_r-\theta_{r-1}\Big\rVert^2\right)
\end{align*}
By incorporating Lemma~\ref{lem:high_order_HAPG} into the inequality above, we have
\begin{align}
\sum_{r=0}^{R-1} \e_r \leq & \left(1-\frac{8 \beta}{9}\right) \sum_{r=-1}^{R-2} \e_r+\frac{8(\lambda \widetilde{L_4})^2}{N K} \mathbb{E}\left[\sum_{r=0}^{R-2}\left\|\nabla J\left(\theta_r\right)\right\|^2\right]+\frac{2 \beta^2 \sigma^2}{N K} R \notag\\
& +\frac{2L^2 +4\widetilde{L_4}^2}{\beta}  8(\eta K)^2\left(2(\eta K L)^2+\frac{1}{K}\right)\left(R \beta^2 \sigma^2+2 L^2 \sum_{r=0}^{R-1} \mathbb{E}\left[\left\|\theta_r-\theta_{r-1}\right\|^2\right]\right) \notag\\
&+\frac{2L^2 +4\widetilde{L_4}^2}{\beta}4eK^2\left\{\frac{\beta^2}{288 e K^2 \left(2L^2+\widetilde{4L_4}^2\right)} \sum_{r=-1}^{R-2}\left(\e_r+\mathbb{E}\left[\left\|\nabla J\left(\theta_r\right)\right\|^2\right]\right)+4 \eta^2 \beta^2 e R G_0 \right\} \notag\\
&\le \left[1-\frac{8 \beta}{9} +\frac{\beta}{72}+\frac{32(\eta K)^2(2L^2+\widetilde{4L_4})^2}{\beta}(2(\eta K L)^2+\frac{1}{K})(\lambda L)^2\right] \sum_{r=-1}^{R-2} \e_r\notag\\
&+\left[\frac{8(\lambda \widetilde{L_4})^2}{N K} + \frac{32(\eta K)^2(2L^2+\widetilde{4L_4})^2}{\beta}(2(\eta K L)^2+\frac{1}{K})(\lambda L)^2 + \frac{\beta}{72}\right]\sum_{r=-1}^{R-2}\mathbb{E}\left[\left\|\nabla J\left(\theta_r\right)\right\|^2\right] \notag\\
&+\left[8\beta\left(2L^2+\widetilde{4L_4}^2\right)(\eta K)^2(2(\eta K L)^2+\frac{1}{K})+\frac{2\beta^2}{NK}\right] R\sigma^2 + 16\beta\left(2L^2+\widetilde{4L_4}^2\right)(e\eta K)^2 R G_0
\end{align}
Where the last inequality is derived by $\left\|\theta_r-\theta_{r-1}\right\|^2 \leq 2 \lambda^2\left(\left\|\nabla J\left(\theta_{r-1}\right)\right\|^2+\left\|u_r-\nabla J\left(\theta_{r-1}\right)\right\|^2\right)$. Note that $\hat{L}^2=2L^2+\widetilde{4L_4}^2.$ We require the following inequalities to hold, 
\begin{equation}
\begin{cases}
&\frac{32(\eta K\hat{L})^2}{\beta}(2(\eta K L)^2+\frac{1}{K})(\lambda L)^2 \le \frac{\beta}{18}\\
&8\hat{L}^2 (\eta K)^2(2(\eta K L)^2+\frac{1}{K}) \le \frac{\beta^2}{NK}\\
& \lambda \hat{L} \le \sqrt{\frac{\beta NK}{72}}.
    \end{cases}  
\end{equation}
Then, we have that 
\begin{align*}
\sum_{r=0}^{R-1} \e_r  &\le \left[1-\frac{8 \beta}{9} +\frac{\beta}{72}+\frac{\beta}{18}\right] \sum_{r=-1}^{R-2} \e_r
+\left[\frac{\beta}{9} + \frac{\beta}{18} + \frac{\beta}{72}\right]\sum_{r=-1}^{R-2}\mathbb{E}\left[\left\|\nabla J\left(\theta_r\right)\right\|^2\right] \notag\\
&+\left[\frac{\beta^2}{NK}+\frac{2\beta^2}{NK}\right] R\sigma^2 + 16\beta(e\eta K \widetilde{L_1})^2 R G_0 \\
& \le (1-\frac{7 \beta}{9})\sum_{r=-1}^{R-2} \e_r + \frac{2\beta}{9}\sum_{r=-1}^{R-2}\mathbb{E}\left[\left\|\nabla J\left(\theta_r\right)\right\|^2\right] +\frac{4R\beta^2\sigma^2}{NK} + 16\beta(e \eta K \hat{L})^2 R G_0 
\end{align*}
After some rearrangement, we have
\begin{align*}
\sum_{r=0}^{R-1} \e_r  &\le \frac{9}{7\beta} \e_{-1} +\frac{2}{7}\sum_{r=-1}^{R-2}\mathbb{E}\left[\left\|\nabla J\left(\theta_r\right)\right\|^2\right] +\frac{36 R\beta \sigma^2}{7NK} +\frac{144}{7}(e\eta K \hat{L})^2 R G_0 
\end{align*}
Based on Lemma~\ref{lem:function_decay}, we have
\begin{align*}
\frac{1}{\lambda}\E[J(\theta_R) -J(\theta_0)] \ge \frac{2}{7}\sum_{r=0}^{R-1}\mathbb{E}\left[\left\|\nabla J\left(\theta_r\right)\right\|^2\right] -\frac{1}{35\beta}\e_{-1}-\frac{39 R\beta \sigma^2}{14NK} -\frac{78}{7}(e\eta K \hat{L})^2 R G_0 
\end{align*}
Notice that $u_0=\frac{1}{N B} \sum_i \sum_{b=1}^B g_i\left(\tau_b^{(i)}|\theta_0\right)$ implies $\e_{-1}=\E\| u_0 -\nabla J(\theta_0)\|^2 \leq \frac{\sigma^2}{N B} \leq \frac{\beta^2 \sigma^2 R}{N K}$. After some rearrangement, we have
\begin{align}
\frac{1}{R} \sum_{r=0}^{R-1} \mathbb{E}\left[\left\|\nabla J\left(\theta_r\right)\right\|^2\right] & \lesssim \frac{\hat{L} \Delta}{\lambda \hat{L} R}+\frac{\e_{-1}}{\beta R}+(\eta K \hat{L})^2 G_0+\frac{\beta \sigma^2}{N K} \notag\\
& \stackrel{(a)}{\lesssim} \frac{\hat{L} \Delta}{\lambda \hat{L} R}+\frac{\beta \sigma^2}{N K} \notag\\
& \stackrel{(b)}{\lesssim} \frac{\hat{L} \Delta}{R}+\frac{\hat{L} \Delta}{\sqrt{\beta N K}}+\frac{\beta \sigma^2}{N K} \notag\\
&\stackrel{(c)}{\lesssim} \frac{\hat{L} \Delta}{R}+\left(\frac{\hat{L} \Delta \sigma}{N K R}\right)^{2 / 3}\notag
\end{align}
where $(a)$ is due to the fact $\eta K \hat{L} \lesssim \left(\frac{\hat{L}\Delta}{G_0 \lambda \hat{L} R}\right)^{\frac{1}{2}};$ For (b), it holds because $\lambda \hat{L} \le \min\{\frac{1}{24}, \sqrt{\frac{\beta NK}{72}}\};$ For (c), it holds because $\beta=\min \left\{1,\left(\frac{N K \hat{L}^2 \Delta^2}{\sigma^4 R^2}\right)^{1 / 3}\right\}.$
\end{proof}

\newpage

\section{Additional Experiments and Implementation Details}

\subsection{Details of Tabular Case.}
Random MDPs consist of $N=20$ environments. In each MDP, both the state and action spaces have a size of $5$. We choose $R_{\max}=1$. The discounted factor $\lambda$ is $0.9$. The state transition kernel is generated randomly (element-wisely Bernoulli distributed). The number of local updates is set as $K=32$. Additionally, the local step-size is chosen to be $\eta =0.05$.

\subsection{Details of DRL Case}

\paragraph{Experiments Setup} We adopted a local step-size of 0.75 and a global step-size of 0.6. We experimented with momentum coefficients, denoted as $\beta$, ranging from 0.2, 0.5, to 0.8. Additional parameters were set as follows: $N = 5$, $R_{\max} = 120$, and $K = 10$. All experiments are conducted in a host machine that is equipped with an Intel(R) Core(TM) i9-10900X CPU that operates at a base frequency of 3.70GHz. This processor boasts 10 cores and 20 threads, with a maximum turbo frequency of 4300 MHz. It has a total of 125GB of RAMA and 4 NVIDIA GeForce RTX 2080 GPU, compatible with CUDA Version 11.0. The source code is provided in the supplementary materials.

\paragraph{Experimental Environments} 

The \textbf{CartPole} environment, often referred to as the "inverted pendulum" problem, is a classic task in the field of reinforcement learning.   In this environment, a pole is attached to a cart, which moves along a frictionless track. The primary objective is to balance the pole upright by moving the cart left or right, without the pole falling over or the cart moving too far off the track.   At the start of the experiment, the pole is slightly tilted, and the goal is to prevent it from falling over by applying force to the cart.   The environment provides a reward at each time step for keeping the pole upright.   The episode ends when the pole tilts beyond a certain angle from the vertical or the cart moves out of a defined boundary on the track. 

The \textbf{HalfCheetah} environment is another popular benchmark in reinforcement learning, especially within the continuous control domain.  It's designed to emulate the challenges of agile and efficient locomotion. The agent in this environment is a two-dimensional, simplified robotic model inspired by the anatomy of a cheetah, albeit it only represents the "half" body, often from the waist down, thus the name "HalfCheetah." The robotic agent comprises multiple joints and segments, representing the limbs of the cheetah.  The primary goal in the HalfCheetah environment is to control and coordinate the movements of these joints to make the robot run as fast as possible on a flat surface. At each timestep, the agent receives a reward based on how fast it's moving forward minus a small cost for the actions taken (to prevent erratic behaviors).  The challenge lies in efficiently propelling the HalfCheetah forward, optimizing for speed and stability. 

The \textbf{Walker} environment is a more complex task that simulates a bipedal agent which needs to learn to walk.   Unlike CartPole, where the challenge is to balance a single pole, the Walker environment involves controlling multiple joints and limbs of a simulated agent to achieve locomotion.   The agent receives rewards based on its forward movement and is penalized for falling or performing awkward movements. More information about these environments can be found in \citet{todorov2012mujoco}.

\begin{figure}[H]
    \centering
    \includegraphics[width=0.45\textwidth]{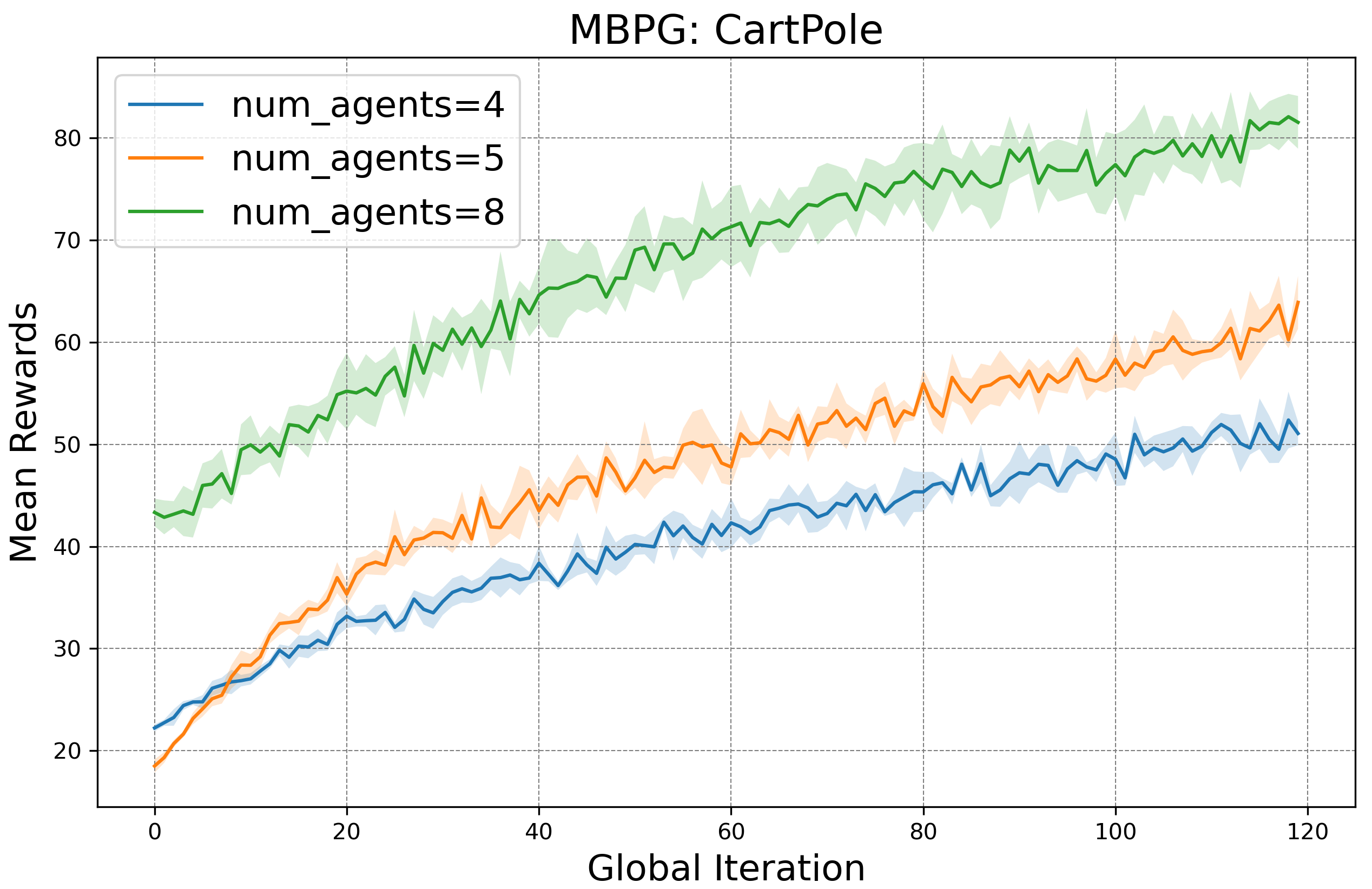}
    \includegraphics[width=0.45\textwidth]{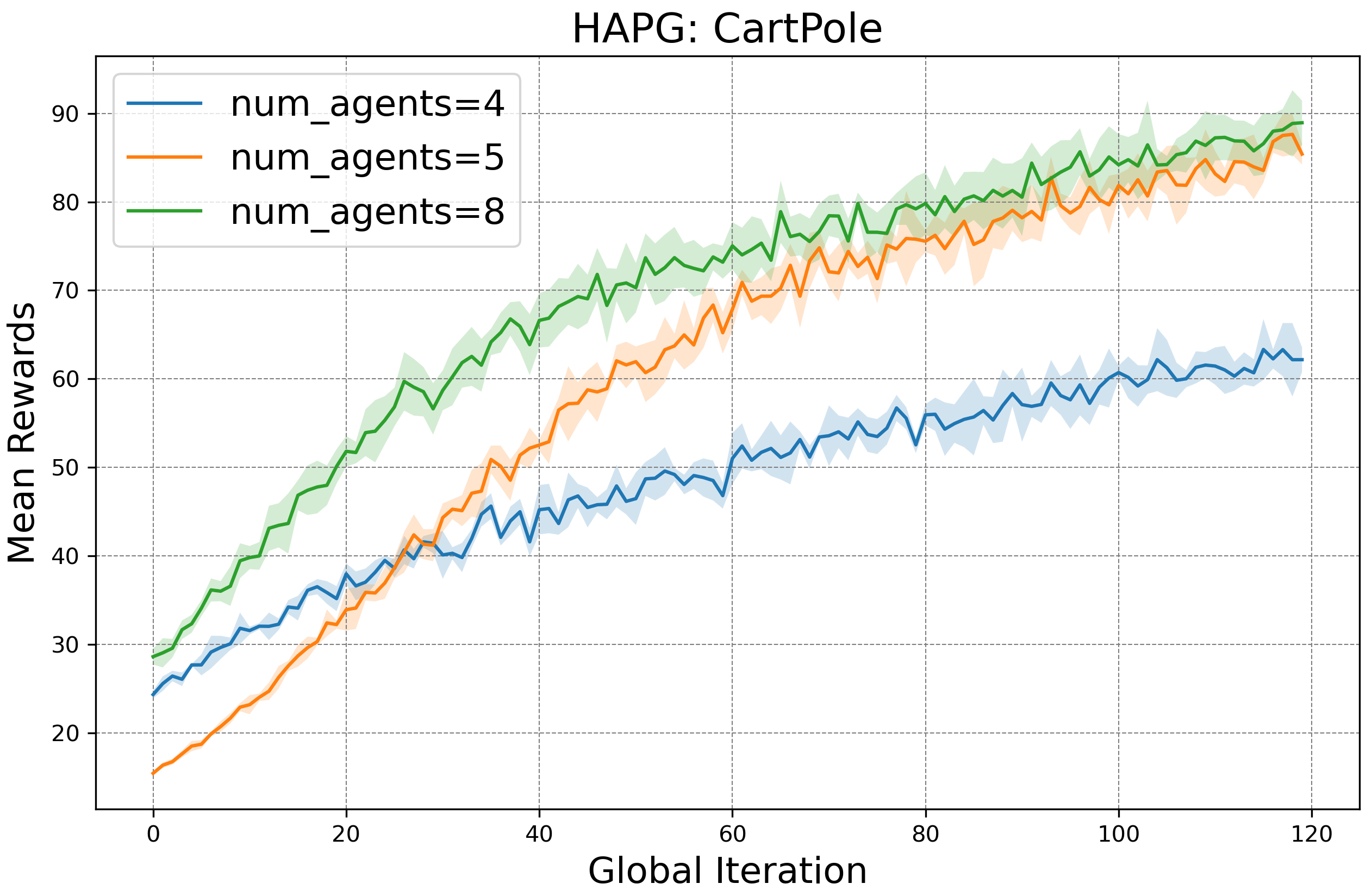}
    \caption{Mean rewards over global iterations for the CartPole task under different values of $N$ (agent number): (\textbf{Left}): \textsc{FedSVRPG-M}; (\textbf{Right}): \textsc{FedHAPG-M}. The shaded areas represent the variance of rewards. Complying with theory, increasing N will increase the rewards. For both algorithms, the local step-size $\eta$ is $0.05$, global step-size $\lambda$ satisfies $\lambda = \eta K$ and the number of local updates $K$ is $10.$} 
    \label{fig:deep_rl_agent_num}
\end{figure}

\paragraph{Ablation Study on Agent Number $N$.} We further provide the {ablation study} of our \textsc{FedSVRPG-M} and \textsc{FedHAPG-M} algorithms on $N$ (agent number). With large $N$, environment heterogeneity level increases. We choose $\beta =0.2$ to train policies in the ablation study. Figure \ref{fig:deep_rl_agent_num} illustrates how different $N$ values ($N=4, 5$, and $8$) 
influence the average rewards in the CartPole task as the number of iterations increases. We find that all policies with larger $N$ values report better performance throughout the iterations. The color-shaded regions indicate the variance in rewards. Such phenomenon observed in Figure \ref{fig:deep_rl_agent_num} complies with our theoretical analysis about linear speedup. 


\paragraph{Experiments on \textsc{FedHAPG-M} Algorihtm}

The table \ref{tab_hapg} presents the mean testing rewards and variances for the policies trained by the FedHAPG-M algorithm with various $\beta$ values and the baseline algorithm \citep{jin2022federated} across two tasks: CartPole and Walker. For both tasks, the FedHAPG-M algorithm with $\beta=0.8$ outperforms the other configurations in terms of mean rewards.

\begin{table}[H]
    \centering
    \caption{Mean Rewards and Variances of Policy Trained by \textsc{FedHAPG-M} with Different Beta Values and Baseline Algorithm}
    \begin{tabular}{c|c|c}
        \toprule
        \text{Algorithms} & \( \text{CartPole} \) & \( \text{Walker} \) \\
        \midrule
         \textsc{FedHAPG-M} with $\beta=0.2$ & \( 83.46 \pm 7.92 \) & \( 130.93 \pm 7.72 \) \\
        \textsc{FedHAPG-M} with $\beta=0.5$ & \( 86.54 \pm 12.99 \) & \( 287.14 \pm 72.26 \) \\
        \textsc{FedHAPG-M} with $\beta=0.8$ & \( \textbf{86.58} \pm 11.21 \) & \( \textbf{301.57} \pm 28.04 \) \\
        \text{Baseline algorithm} & \( 85.92 \pm 12.17 \) & \( 299.69 \pm 3.02 \) \\
        \bottomrule
    \end{tabular}
    \label{tab_hapg}
\end{table}


\end{document}